\documentclass[11pt]{article}
\usepackage{}
\usepackage{amsfonts}
\usepackage{mathrsfs}
\usepackage{amssymb}
\usepackage{amsmath}
\usepackage{bbm}
\usepackage{cite}
\usepackage{color}
\usepackage{extarrows}
\usepackage{graphicx}
\usepackage{subfigure}
\usepackage{float}
\usepackage{diagbox}
\usepackage{booktabs}
\usepackage{algorithm}
\usepackage{algpseudocode}
\usepackage{caption}
\usepackage{color}
\usepackage{xcolor}

\allowdisplaybreaks[4]

\def\no{\nonumber}

\newcommand\btd{\raise 2pt \hbox{$\hat\bigtriangledown$}\hskip 1.5pt}
\newcommand\bt{\raise 2pt \hbox{$\bigtriangledown$}\hskip 1.5pt}

\def\no{\nonumber}

\begin{document}
\title{Symmetry group based domain decomposition to enhance physics-informed neural networks for solving partial differential equations}
\author{ Ye Liu $^1$\ \ \ Jie-Ying Li $^1$\ \ \ Li-Sheng Zhang $^2$ \ \ \  Lei-Lei  Guo $^2$\ \ \ Zhi-Yong Zhang $^1$\footnote{E-mail: zzy@muc.edu.cn}
 \\
\small $^1$ College of Science, Minzu University of China, Beijing 100081, P.R. China\\
\small $^2$ College of Science, North China University of Technology, Beijing 100144, P.R. China}
\date{}
\maketitle

\noindent{\bf Abstract:} Domain decomposition provides an effective way to tackle the dilemma of physics-informed neural networks (PINN) which struggle to accurately and efficiently solve partial differential equations (PDEs) in the whole domain, but the lack of efficient tools for dealing with the interfaces between two adjacent sub-domains heavily hinders the training effects, even leads to the discontinuity of the learned solutions. 
In this paper, we propose a symmetry group based domain decomposition strategy to enhance the PINN for solving the forward and inverse problems of the PDEs possessing a Lie symmetry group. Specifically, for the forward problem, we first deploy the symmetry group to generate the dividing-lines having known solution information which can be adjusted flexibly and are used to divide the whole training domain into a finite number of non-overlapping sub-domains, then utilize the PINN and the symmetry-enhanced PINN methods to learn the solutions in each sub-domain and finally stitch them to the overall solution of PDEs. For the inverse problem, we first utilize the symmetry group acting on the data of the initial and boundary conditions to generate labeled data in the interior domain of PDEs and then find the undetermined parameters as well as the solution by only training the neural networks in a sub-domain. Consequently, the proposed method can predict high-accuracy solutions of PDEs which are failed by the vanilla PINN in the whole domain and the extended physics-informed neural network in the same sub-domains. Numerical results of the Korteweg-de Vries equation with a translation symmetry and the nonlinear viscous fluid equation with a scaling symmetry show that the accuracies of the learned solutions are improved largely.


\section{Introduction}
Partial differential equations (PDEs) are the most widely used models to describe the natural phenomenon and scientific problems, thus finding solutions of PDEs is an important but intractable subject. Since constructing analytical solutions is not unrealistic for most PDEs, numerical computation methods, such as finite difference \cite{ran-2007} and finite element \cite{ole-2005}, etc. have received a great deal of attentions as a practical technique to understand complex phenomena that are
almost impossible to be treated analytically. For example, the finite element method can effectively solve the problem of Stokes flow in porous media but still suffers from a highly non-trivial task in terms of human efforts and the actual computational cost \cite{my-2021}. And the dimension constraints and feasibility for solving high-dimensional PDEs with such methods is also an unbreakable issue.

In recent years, with the improvement of computational power and the innovation of algorithms, using neural networks to solve PDEs has gained a lot of popularity \cite{han-2018,2018a}. The idea can be traced back to Lee and Kang in the $1990s$ who utilized neural networks as trial functions in PDE solvers \cite{lee-1990}, where the mathematical foundation was the universal approximation theorems stating that a single layer neural network can approximate most
functions in Sobolev spaces \cite{km-1989}. Later, Lagaris et.al utilized the hard constraint technique for the trial solution of a differential equation which was written as a sum of the known initial/boundary conditions without any adjustable parameters and a feed-forward neural network containing adjustable parameters \cite{lia-1998}. Consequently, the deep Ritz method \cite{e-2018} and the deep Galerkin method \cite{gm-2018} were proposed to study the variational problems arising from the ordinary PDEs and the high-dimensional PDEs, respectively. In particular, Raissi et al. proposed the physics-informed neural networks (PINN) which transform the problem of inferring solutions of PDEs into an optimization problem of the loss function which consists of PDE residual together with the initial/boundary conditions constraints \cite{2018a}. Up to now, by analyzing the pathologies of the PINN in failure scenarios, some improved versions have been proposed and widely applied in the field of scientific computing, especially in solving forward and inverse problems of nonlinear PDEs due to the merits of flexibility and gridless nature \cite{wang-2021,zhang-2023,lmz-2021}.

Despite the great efforts and marvelous results, the current situation for solving PDEs via the PINN is not entirely satisfactory where the limited accuracy is one of the most concerns. Currently, the PINN method usually achieves accuracies from $10^{-2}$ to $10^{-3}$ if it works, but often fails to predict certain particular solutions of PDEs such as the large amplitude and high-frequency solutions \cite{wang-2021,zhang-2023} as well as the solutions of PDEs with large diffusion coefficient \cite{ak-2021}. Therefore, several effective techniques were proposed and incompletely summarized as follows according to the components of the training neural network.

$\bullet$ \emph{Loss function}. The first direct and effective tool is to design an appropriate loss function to optimize the quantities of interest. For example, a gradient-enhanced PINN was proposed in solving both forward and inverse problems of PDEs where the gradient information of the PDEs residual was embedded into the loss function \cite{J}. Lin and Chen introduced a two-stage PINN where the first stage is the standard PINN and the second stage is to incorporate the conserved quantities of PDEs into mean squared error loss to train the neural networks \cite{jan}. In \cite{tb-2022}, the authors enforced nonlinear analytic constraints into the loss function to produce results consistent with the constraints. Recently, we added the invariant surface condition induced by the Lie symmetry of PDEs into the loss function of PINN to improve the efficiencies for the forward and inverse problems of PDEs \cite{zhang-2022}.

$\bullet$ \emph{Architecture of neural network}. The second one is to insert the physical properties into the architecture of neural networks, which seems harder to be performed than to enhance the constraints of the loss function. The celebrated convolutional neural network keeps invariance along the groups of symmetries \cite{ma-2016}, while the covariant neural network conforms with the rotation and translation invariance in many-body molecular systems \cite{kon-2018}. Recently, Zhu et.al proposed a group-equivariant neural network method which respects the spatio-temporal parity symmetries and successfully emulates different types of periodic solutions of nonlinear dynamical lattices \cite{zhu-2022}. Beucler et.al incorporated the nonlinear analytic constraints as a constraint layer in the architecture of neural networks to make the predictions exactly keep the physical properties of PDEs except for the machine errors \cite{tb-2022}.

$\bullet$ \emph{Training domain}. Since the predicted solutions in different domains may have different properties, thus dividing the total domain into several sub-domains may improve the accuracies of the learned solution. In fact, the method of domain decomposition is widely used in numerical simulation for solving PDEs related to the problems of fluid flow \cite{tang-2020}, wave propagation, and heat flow \cite{vd-2015}, which inspires the technique in the field of machine learning. For example, Jagtap et al. proposed a conservative physics-informed neural network on sub-domains for nonlinear conservation laws where the continuity of the states and their fluxes across sub-domain interfaces are deployed to stitch the sub-domains together \cite{jag-2020a}. Later,
Jagtap and Karniadakis relied on the domain decomposition technique to propose the extended PINN (XPINN) for any type of PDEs which offers both space-time domain decomposition for any irregular, non-convex geometry thereby reducing the computational cost effectively \cite{jag-2020b}. In \cite{elh-2023}, the original XPINN method was augmented by imposing flux continuity across the domain interfaces.
The domain decomposition technique is applied to the variational formulation of PINN \cite{kha-2021a} and the time-dependent PDEs where a long-time problem was decomposed into many independent short-time problems \cite{meng-2020,shu-2021}.
Hence, domain decomposition provides an opportunity to employ individual networks in each sub-domain, and then the overall solution is reconstructed by stitching
the learned solutions by individual networks in each sub-domain. A key consideration in any domain decomposition approach is to ensure that the individual sub-domain
solutions are communicated across the sub-domain interfaces, which relies upon additional interface terms in their PINN loss function. Therefore, it requires special treatment at the interfaces of two adjacent sub-domains which is the biggest barrier of the domain decomposition method \cite{jag-2020a}.

Our objective in this paper is to propose a symmetry group based domain decomposition for PINN (sdPINN) to solve the forward and inverse problems of PDEs that shares the merits of the symmetry group of PDEs, domain decomposition and PINN method. Specifically, for the forward problem, we first leverage the invariants of the symmetry group of PDEs and the discrete points on the known initial and boundary conditions to generate the dividing lines which separate the whole domain into a finite number of non-overlapping sub-domains, then in each sub-domain perform PINN or sPINN to learn the solution of PDEs.
The distinct merit of the proposed method is the interfaces of two adjacent sub-domains which have the exact labeled data, and thus in each sub-domain one can perform completely different neural networks including the choice of different network architecture, number of training points and activation functions, optimization algorithm, etc. Consequently, the training works in a highly parallel fashion and thus saves the training times. Such a scene of completely independent training in each sub-domain makes that the training for the inverse problem of PDEs can be performed in a sub-domain, not the whole domain of PDEs. Moreover, the required labeled data for the inverse problem can be obtained by mapping the data on initial and boundary conditions to generate labeled data in the interior domain with the Lie symmetry group.
The key point of the introduced method is the symmetry group which endows each sub-domain the exact initial and boundary conditions and thus makes the training in each sub-domain completely independent. Thus the idea used in this paper is a direct and specific application of the Lie symmetry group keeping the PDEs invariant rather than its conventional symmetry reduction ability for PDEs, while it is also an innovative thinking of extending Lie symmetry theory to the numerical solution field of PDEs and may provide some potential inspiration for further researches on this issue.

Of course, it is observed that there exists a constraint of the proposed method that the PDEs under study must be admitted by a symmetry group. In fact, symmetry is an inherent and connotative characteristic of PDEs and many PDEs possess usual symmetry groups of physical meanings \cite{blu,olv}. For example, a linear homogeneous PDEs always admits the scaling transformation group $ t^*=t, x^*=x, u^*=\lambda u$ with a scaling constant $\lambda$, while a translation group of time and space, i.e. $ t^*=t+\lambda\epsilon, x^*=x+\mu\epsilon, u^*=u$ with a group parameter $\epsilon$ and the two constants $\lambda$ and $\mu$ controlling the translation sizes, always leaves invariant the PDEs in which the independent variables $t$ and $x$ do not appear explicitly. Furthermore, there exist some complex symmetry groups which are hard to be detected by observation but can be obtained by the known algorithms \cite{olv}. Thus, starting from the symmetry group of PDE to enhance the learning and generalization abilities of PINN has a solid foundation.

The remainder of the paper is outlined as follows. In Section 2, we take the Korteweg-de Vries (KdV) equation as an example to recall the main idea of the vanilla PINN and the XPINN, and further state the problem formulation for learning a large amplitude solution of the KdV equation. In Section 3, we propose a symmetry group based domain decomposition to enhance the PINN to solve the forward problem of PDEs and show the effectiveness of the introduced method by performing two numerical experiments including the KdV equation and the nonlinear viscous fluid equation. In Section 4, we use the introduced method to solve the inverse problem of PDEs, where, in particular, the whole training only works in a sub-domain of the forward problem and the required labeled data in the interior domain are uniquely generated by acting the symmetry group on the discrete data of initial and boundary conditions. We conclude the results in the last section.
\section{Problem formulation}
Consider the KdV equation
\begin{eqnarray} \label{KdV}
&& u_t+uu_{x}+u_{xxx}=\mu(x,t),~~~~(x,t)\in[-1,1]\times[0,1],
\end{eqnarray}
which arises in the theory of long waves in shallow water and the physical systems in which both nonlinear and dispersive effects are relevant \cite{KdV}. We want to utilize the PINN method to learn the solution $u_{kdv}=\left(x-2t\right)^2+b\,\sin(\pi(x-2t))$, where $b$ is a nonzero constant, the initial and boundary conditions as well as the function $\mu=\mu(x,t)$ in Eq.(\ref{KdV}) are enforced by the solution $u_{kdv}$.

The PINN method is a feed-forward neural network which incorporates the residuals of the governing PDEs and the initial and boundary conditions as the loss function, i.e.
\begin{eqnarray} \label{PINN}
&& \mathcal {L}(\theta)=MSE_u+MSE_f,
\end{eqnarray}
where the mean square errors of initial and boundary conditions $MSE_u$ and the one for Eq.(\ref{KdV}) $MSE_f$ are defined by
\begin{eqnarray} \label{PINN-MSE}
&&\no MSE_u=\frac{1}{N_u}\sum_{i=1}^{N_u}\bigg[~{\mid u(x_{u}^i, 0)-(x_{u}^i)^2-b\sin({\pi}x_{u}^i)\mid}^{2}\\
&&\no\hspace{3.2cm}+{\mid u(-1, t_{u}^i)-(1+2t_{u}^i)^2+b\sin({\pi}(1+2t_{u}^i))\mid}^{2}\\
&&\no\hspace{3.2cm}+{\mid u(1, t_{u}^i)-(1-2t_{u}^i)^2-b\sin({\pi}(1-2t_{u}^i))\mid}^{2}~\bigg],\\
&&\no MSE_f=\frac{1}{N_f}\sum_{i=1}^{N_f}{\mid f(x_{f}^i, t_{f}^i)\mid}^{2},
\end{eqnarray}
where $\left\{(x_{u}^{i},t_{u}^{i},u(x_{u}^{i},t_{u}^{i}))\right\}_{i=1}^{N{_u}}$ denote the initial and boundary data and $\big\{(x_{f}^{i},t_{f}^{i})\big\}_{i=1}^{N_f}$ stand for the collocations points for $f(x,t):=u_t+uu_{x}+u_{xxx}-\mu$. To obtain the training data, we discretize the spatial domain $x\in[-1,1]$ into $N_x=400$ and the temporal domain $t\in[0,1]$ into $N_t=200$ discrete equidistant points. The solution $u_{kdv}$ is then discretized into $400\times200$ data points in the given space-time domain $[-1,1]\times[0,1]$.
In the experiments, we use a 4-layer fully connected neural network with 40 neurons per layer, a hyperbolic tangent activation function and the L-BFGS optimizer \cite{ln-1989} to minimize the loss function (\ref{PINN}). Moreover, the weight matrixes and bias vectors are initialized by the Xavier initialization \cite{xw-2010} and the derivatives of $u(x,t)$ with respect to time $t$ and space $x$ are computed by automatic differentiation \cite{auta}.

\begin{figure}[htp]
    \begin{center}
	\begin{minipage}{0.46\linewidth}
		\vspace{3pt}
		\centerline{\includegraphics[width=\textwidth]{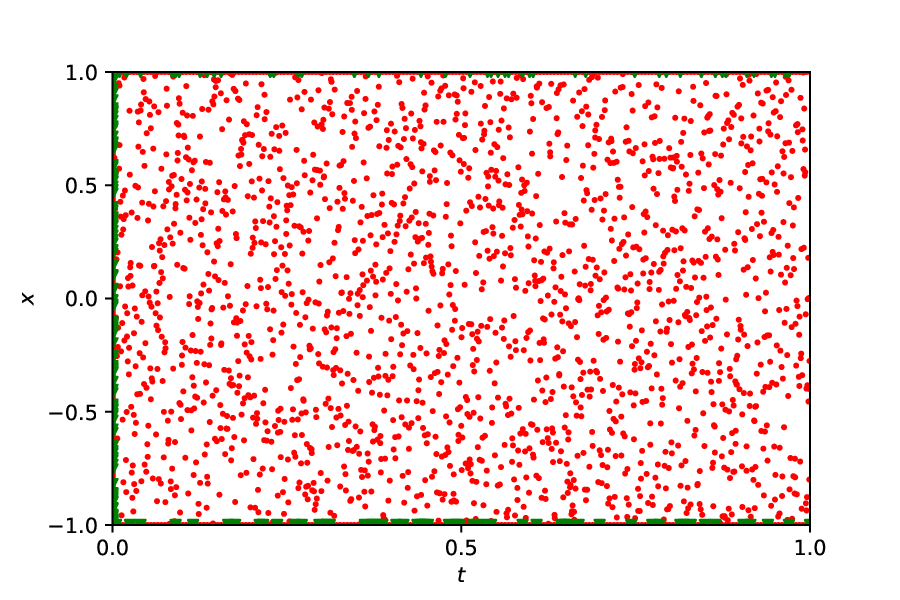}}
        \centerline{A}
	\end{minipage}
    \begin{minipage}{0.475\linewidth}
		\vspace{3pt}
		\centerline{\includegraphics[width=\textwidth]{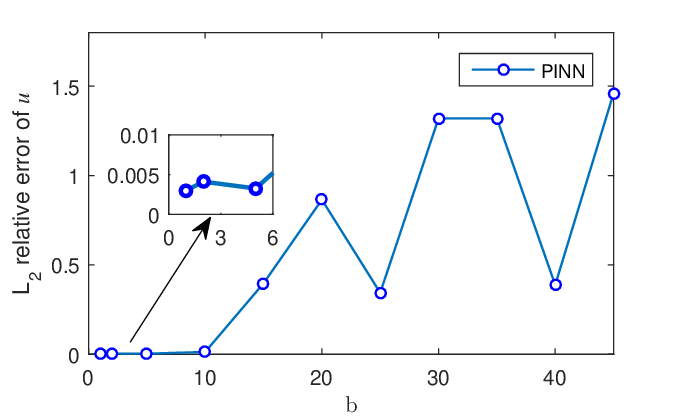}}
        \centerline{B}
	\end{minipage}
    \end{center}
	\caption{(Color online) KdV equation: (A) Distribution of training data for the PINN. (B) $L_2$ relative errors by the PINN with respect to different $b$.}
\label{fig1}
\end{figure}

Figure \ref{fig1}(A) shows the distribution of red collocation points via the Latin hypercube sampling \cite{ms-1987} and the green points which are randomly chosen from the set of the initial and boundary points. Figure \ref{fig1}(B) displays the $L_2$ relative errors of the predicted solutions by the PINN with respect to different $b$. Observed that the PINN method is only able to achieve comparatively good predicted solutions for $b\leq10$ whose the $L_2$ relative errors are less than $10^{-2}$, but for $b>10$ the $L_2$ relative errors increase largely and fluctuate around $10^{-1}$ order of magnitude, which means that the PINN method no longer works for $b>10$.

\begin{figure}[htp]
	\begin{minipage}{0.33\linewidth}
		\vspace{3pt}
		\centerline{\includegraphics[width=\textwidth,height=0.8\textwidth]{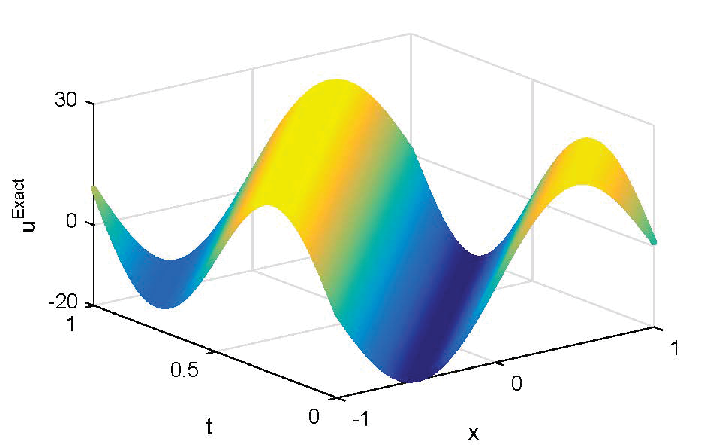}}
        \centerline{A}
	\end{minipage}
    \begin{minipage}{0.33\linewidth}
		\vspace{3pt}
		\centerline{\includegraphics[width=\textwidth,height=0.8\textwidth]{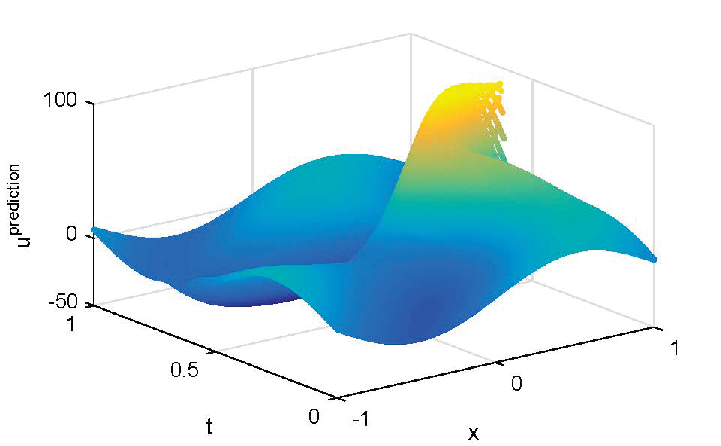}}
        \centerline{B}
    \end{minipage}
    \begin{minipage}{0.33\linewidth}
		\vspace{3pt}
		\centerline{\includegraphics[width=\textwidth,height=0.8\textwidth]{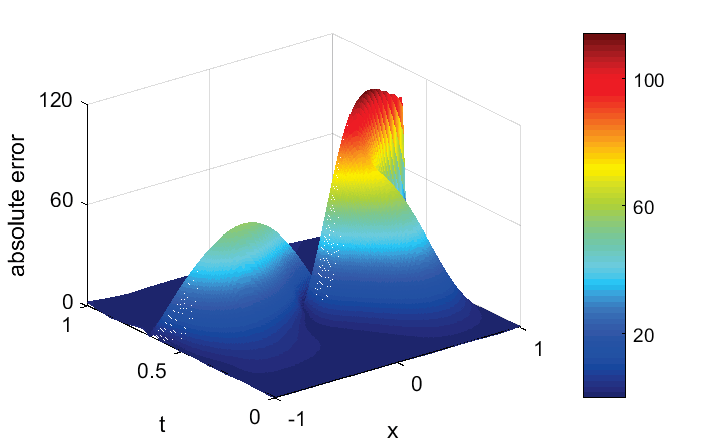}}
        \centerline{C}
	\end{minipage}
	\caption{(Color online) KdV equation: comparisons of exact solution and predicted solution and absolute errors of PINN. (A) Exact solution. (B) Predicted solution. (C) Absolute errors.}
\label{fig2}
\end{figure}

In particular, we choose a special case of solution $u_{kdv}$, i.e. $b= 20$, to show the prediction of PINN against the exact solution of Eq.(\ref{KdV}) and display the visual
comparisons of the exact solution and predicted solution in Figure \ref{fig2}(A-B) respectively. The absolute errors in Figure \ref{fig2}(C) show that the PINN method does a poor job in learning the solution $u_{kdv}$ with $b=20$, leading to the maximum absolute error value $114.2423$.

\section{Forward problem}
Consider the following $r$-th order PDE
\begin{eqnarray} \label{eqn1}
&& f: =u_t+\mathcal {N}[u]=0,~~~~~t\in [0, T],~~x\in [a,b],
\end{eqnarray}
together with the initial and boundary conditions
\begin{eqnarray}\label{ib}
&&\no u(x,0)=\phi(x),\\
&& u(a,t)=\varphi_a(t),~~~~u(b,t)=\varphi_b(t),
\end{eqnarray}
where $u=u(x, t)$ is the solution to be determined, $\phi(x)$ is a smooth function in $[a,b]$, $\varphi_a(t)$ and $\varphi_b(t)$ denote two smooth functions in $[0,T]$, and $\mathcal{N}[u]$ denotes a smooth function of $u$ and its $x$-derivatives up to $r$th order.
\subsection{Lie symmetry}
Lie symmetry is an inherent but not exposed property of PDEs. 
The classical method for obtaining a Lie symmetry admitted by Eq.(\ref{eqn1}) is to find a local one-parameter transformation group
\begin{eqnarray}\label{group}
&&\no x^*= x+\varepsilon\,\xi (x,t,u)+O(\varepsilon^2),\\
&&\no t^*=t+\varepsilon\,\tau (x,t,u)+O(\varepsilon^2),\\
&& u^*=u+\varepsilon\,\eta (x,t,u)+O(\varepsilon^2),
\end{eqnarray}
which leaves Eq.(\ref{eqn1}) invariant. Lie's fundamental theorem shows that such a group (\ref{group}) is completely characterized by the
infinitesimal operator
\begin{eqnarray}\label{oper}
&& \mathcal {X}=\xi\partial_x+\tau\partial_t+\eta\partial_u,
\end{eqnarray}
where the infinitesimals $\xi=\xi(x,t,u),\tau=\tau(x,t,u)$ and $\eta=\eta(x,t,u)$, thus we will not differentiate the Lie group (\ref{group}) and the corresponding operator $\mathcal {X}$ and call them as Lie symmetry.
Lie's infinitesimal criterion for Eq.(\ref{eqn1}) requires $\mathcal {X}$ satisfying $\text{pr}^{(r)}\mathcal {X}(f)_{|\{(\ref{eqn1})\}}=0$ \cite{olv,blu},
where the symbol $_{|\{\Delta\}}$ means that the computations work on the solution space of $\Delta=0$, and
 \begin{eqnarray}
&&\no \text{pr}^{(r)}\mathcal {X}=\mathcal {X}+\eta_t^{(1)}\partial_{u_t}+\eta_x^{(1)}\partial_{u_1}+\eta_x^{(2)}\partial_{u_2}+\dots+\eta_x^{(r)}\partial_{u_r}
\end{eqnarray}
stands for $r$-th order prolongation of $\mathcal {X}$, $u_i=\partial^i u/\partial x^i$ with $i=1,2,\dots,r$, the coefficients $\eta_t^{(1)}$ and $\eta_x^{(i)}$ can be calculated by the well-known prolongation formulae \cite{blu,olv},
 \begin{eqnarray}\label{formula}
&&\no \eta_t^{(1)}=D_t\left(\eta-\tau u_t-\xi\,u_x\right)+\tau D_t^2u+\xi\,D_xD_tu,\\
&& \eta_x^{(i)}=D_x^i\left(\eta-\tau u_t-\xi\,u_x\right)+\tau D_tD_x^iu+\xi\,D_x^{i+1}u.
\end{eqnarray}
Moreover, the symbols $D_t$ and $D_x$ in (\ref{formula}) indicate the total derivatives with respect to $t$ and $x$ respectively,
 \begin{eqnarray}
&&\no D_t=\partial_t+u_t\partial_u+u_{xt}\partial_{u_{x}}+u_{tt}\partial_{u_t}+\dots,\\
&&\no D_x=\partial_{x}+u_1\partial_u+u_2\partial_{u_1}+u_{xt}\partial_{u_t}+\dots,
\end{eqnarray}
and $D_t^0(u)=u$, {$D_t^{i}=D_t\left(D_t^{i-1}\right)$ }and similar for $D_x$. {Since $\xi,\tau$ and $\eta$ are independent of $x$- and $t$-derivatives of $u$, then on the solution space of Eq.(\ref{eqn1}), by equating the coefficients of various monomials in the different order of $x$- and $t$-derivatives of $u$ to zero, we obtain an over-determined linear differential system whose solutions are the required Lie symmetries. For details we refer to references \cite{olv,blu}.}

Furthermore, the set of all Lie symmetries of Eq.(\ref{eqn1}) forms a Lie algebra of vector field $\mathcal{X}$, which means that a linear combination of Lie symmetry of Eq.(\ref{eqn1}) is also a Lie symmetry of Eq.(\ref{eqn1}). Lie symmetry has wide applications where a primary role of Lie symmetry is to map one solution into another solution of the same PDEs \cite{olv}. For example, the governing KdV equation in Eq.(\ref{KdV}) is admitted by a combination of time and space translation groups, $G_{kdv}: t^*=t+\beta\,\epsilon\,, x^*= x+\alpha\,\epsilon\,, u^*=u$ with a group parameter $\epsilon$ and two nonzero constants $\alpha$ and $\beta$ of translation sizes. Then for a given point $(t_0,x_0,u(x_0,t_0))$ on a solution surface, the symmetry group $G_{kdv}$ transforms it to the point $(t_0+\beta\,\epsilon,x_0+\alpha\,\epsilon,u(x_0,t_0))$ which is still on the same solution surface. Such a peculiar property will be used to generate labeled data in the interior domain of PDEs from the ones on the initial and boundary conditions.

Another main effect of Lie symmetry is to reduce the PDEs into low-dimensional PDEs by means of the invariants of $\mathcal{X}$. The invariant $I=I(x,t,u)$ of $\mathcal {X}$ is determined by $\mathcal {X}(I)=\xi\partial_x(I)+\tau\partial_t(I)+\eta\partial_u(I)=0$. Then solving the corresponding characteristic equations $dx/\xi=dt/\tau=du/\eta$ gives two invariants $I_1(x,t,u)=C_1$ and $I_2(x,t,u)=C_2$, where $C_1$ and $C_2$ are two integrated constants. For example, the two invariants of Lie symmetry $G_{kdv}$ of Eq.(\ref{KdV}) are $I_1=\alpha t-\beta x$ and $I_2=u$ by solving $\alpha\partial_x(I)+\beta\partial_t(I)=0$. Such invariants of $\mathcal{X}$ will be used to find the dividing lines in the domain decomposition method.
\subsection{sdPINN for the forward problem}
We first use the symmetry group (\ref{oper}) to generate exact known interfaces of any two neighboring sub-domains from the initial and boundary conditions (\ref{ib}), and then perform a completely independent PINN or its improved versions to learn the solution of Eq.(\ref{eqn1}) in each sub-domain, and finally stitch them to obtain the solution in the whole domain. 

Specifically, suppose that Eq.(\ref{eqn1}) is admitted by the symmetry $\mathcal {X}$ defined by (\ref{oper}) whose two invariants are $I_1(x,t,u)=C_1$ and $I_2(x,t,u)=C_2$. Then for a discrete point $P(x_{ib},t_{ib},u(x_{ib},t_{ib}))$ on the initial and boundary conditions (\ref{ib}), there exists an integrated constant $C_1$ such that the invariant $I_1(x,t,u)=C_1$ starts with the point $P$ and across the interior domain and ends at the other boundary point. Thus the dividing-line described by the invariant $I_1(x,t,u)=C_1$ divides the whole domain into two sub-domains. More importantly, the discrete points on the dividing-line are known by means of the starting point $P$ and symmetry group (\ref{group}), i.e.
\begin{eqnarray}
\left\{(x_{ib}+\varepsilon\,\xi+O(\varepsilon^2),t_{ib}+\varepsilon\,\tau+O(\varepsilon^2),u(x_{ib},t_{ib})+\varepsilon\,\eta+O(\varepsilon^2))\right\}.
\end{eqnarray}
Furthermore, for other points instead of $P$ on the initial and boundary conditions, one can follow the above step to get the new dividing-lines and thus divide the whole domain into a finite number of non-overlapping sub-domains.

Assume that the number of such non-overlapping sub-domains is $n$. Then in the $p$th sub-domain, we perform PINN method and thus define the loss function as
\begin{eqnarray} \label{sdPINN}
&& \no \mathcal {L}_p(\theta)=w_{u_p} MSE_{u_p}+w_{f_p} MSE_{f_p},
\end{eqnarray}
where, here and after, $w_{u_p}$ and $w_{f_p}$ are the weight parameters, $MSE_{u_p}$ and $MSE_{f_p}$ are
\begin{eqnarray} \label{sdPINN-MSE}
&&\no MSE_{u_p}=\frac{1}{N_{u_p}}\sum_{i=1}^{N_{u_p}}\bigg[~{\mid u(x_{u_p}^i,t_{u_p}^i)-u^i\mid}^{2} + {\mid u_p(x_{I_p}^i,t_{I_p}^i)-u_p^i\mid}^{2}~\bigg],\\
&& MSE_{f_p}=\frac{1}{N_{f_p}}\sum_{j=1}^{N_{f_p}}{\mid f(x_{f_p}^j,t_{f_p}^j)\mid}^{2}.
\end{eqnarray}
Here, $\{(x_{u_p}^{i},t_{u_p}^{i},u^{i})\}_{i=1}^{N_{u_p}}$ denote the initial and boundary training data in (\ref{ib}), $\big\{(x_{I_p}^{i},t_{I_p}^{i})\big\}_{i=1}^{N_{u_p}}$ denotes the training data on the interfaces of sub-domains and $\big\{(x_{f_p}^{j},t_{f_p}^{j})\big\}_{j=1}^{N_{f_p}}$ specify the collocations points for $f(x,t)$.

\begin{figure}[htp]
    \begin{center}
    \begin{minipage}{1\linewidth}
		\vspace{3pt}
		\centerline{\includegraphics[width=\textwidth]{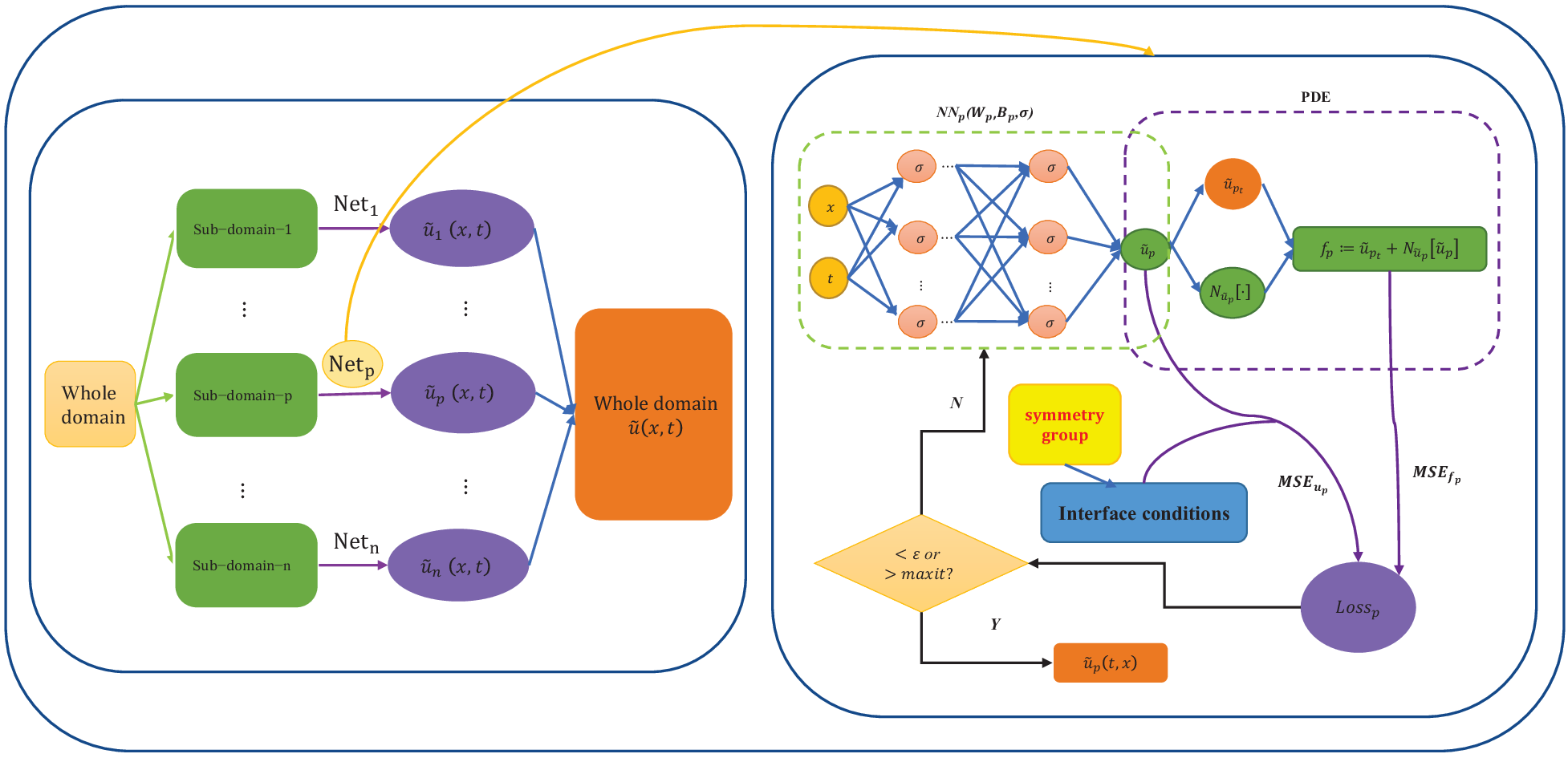}}
	\end{minipage}
    \end{center}
	\caption{(Color online): Schematic diagram of the sdPINN method. }
\label{fig3}
\end{figure}

Figure \ref{fig3} shows the schematic diagram of the sdPINN algorithm. The left sub-figure in Figure \ref{fig3} shows that the sdPINN first divides the whole domain into $n$ sub-domains, and then in each sub-domain one uses a completely different neural network to train, while the right one displays the whole neural network in the $p$th sub-domain where the exact interfaces obtained by the Lie symmetry are added in $MSE_{u_p}$ of the loss function.


Furthermore, since the Lie symmetry (\ref{oper}) of Eq.(\ref{eqn1}) induces the invariant surface condition $g(x,t):=\xi u_x+\tau u_t-\eta=0$ which enhances the constraints on the learned solution, thus we add it into the loss function and call the method as sdPINN-isc. The sdPINN-isc greatly accelerates the convergence speed, improves the computational efficiency and gets high-accuracy predicted solutions \cite{zhang-2023}. Then the loss function of the sdPINN-isc for the $p$th sub-domain is
\begin{eqnarray} \label{invaloss-enh}
&& \mathcal {L}_p(\theta)=w_{u_p} MSE_{u_p}+w_{f_p} MSE_{f_p}+w_{g_p} MSE_{g},
\end{eqnarray}
where $MSE_{u_p}$ and $MSE_{f_p}$ are defined in (\ref{sdPINN-MSE}), $MSE_{g}$ is given by
\begin{eqnarray} \label{invaloss-enh}
&&\no MSE_{g_p}=\frac{1}{N_{g_p}}\sum_{i=1}^{N_{g_p}}{\mid g(x_{g_p}^{i},t_{g_p}^{i})\mid}^{2}.
\end{eqnarray}

\subsection{Comparisons with the extended PINN (XPINN)}
The XPINN incorporates the domain decomposition into the PINN and thus divides the whole domain into several non-overlapping sub-domains in which different PINNs are used. Compared with the conservative PINN in \cite{J}, the major improvement of XPINN is to replace the conservative quantities by the mean solution on the common interface of two adjacent sub-domains which makes the XPINN suitable for solving any type of PDEs theoretically \cite{jag-2020b}.

Specifically, in the $p$th sub-domain, the loss function of XPINN is given by
\begin{eqnarray} \label{xpinn-loss}
&&\no \mathcal{L}_p(\theta)=w_{u_p} MSE_{u_p}+w_{f_p} MSE_{f_p}+w_{I_p} (MSE_{R}+MSE_{u_{avg}}),
\end{eqnarray}
where $MSE_{u_p}$ and $MSE_{f_p}$ are defined by (\ref{sdPINN-MSE}), $MSE_{R}$ and $MSE_{u_{avg}}$ are given by
\begin{eqnarray} \label{xpinn-loss-mse}
&&\no MSE_{R}=\frac{1}{N_{I_p}}\sum_{i=1}^{N_{I_p}}{\mid f_p(x_{I_p}^{i},t_{I_p}^{i})-f_{p^{+}}(x_{I_p}^{i},t_{I_p}^{i})\mid}^{2},\\
&& MSE_{u_{avg}}=\frac{1}{N_{I_p}}\sum_{i=1}^{N_{I_p}}{\mid u(x_{I_p}^{i})-u_{avg}\mid}^{2},
\end{eqnarray}
where $f_p$ is the residual of the governing PDE, the $MSE_R$ is the residual continuity condition on the common interface given by two
different neural networks on two neighboring sub-domains $p$ and ${p}^{+}$, $\big\{(x_{I_p}^{i},t_{I_p}^{i})\big\}_{i=1}^{N_{I_p}}$ denotes the point set on the common interface between the two neighboring sub-domains $p$ and ${p}^{+}$. The $u_{avg}$ is averaged along the common interface as $u_{avg}=(u_p+ u_{{p}^{+}})/{2}$.

Therefore, compared with the XPINN, in addition to the independent computational node in each sub-domain including training points, activation functions, the width and depth of the network, and so on, the main contribution of the sdPINN is the exact interfaces of any two adjacent sub-domains which make the sdPINN has some distinct merits:


$\bullet$ \emph{Keep the learned solution continuous in the whole domain}: In the XPINN, the interfaces of two adjacent sub-domains are constrained by the $MSE_R$ and $MSE_{u_{avg}}$ in (\ref{xpinn-loss-mse}), which sometimes makes the stitched solution in the whole domain discontinuous at the interfaces, see the two numerical experiments in the next subsection. However, the sdPINN has exact interfaces in each two neighboring sub-domains and thus keeps the learned solution continuous in the whole domain.

$\bullet$ \emph{Avoiding error propagation in the domain}: The two terms $MSE_{R}$ and $MSE_{u_{avg}}$ in (\ref{xpinn-loss-mse}) heavily depend on the accuracy of the previous sub-domain and thus the poor accuracy of the previous sub-domain may contaminate the learned solution in the following sub-domain and finally make the learned solution in the whole domain with bad accuracy.


$\bullet$ \emph{Parallelization capability}: In the XPINN, any two sub-domains are directly or indirectly connected by $MSE_{R}$ and $MSE_{u_{avg}}$ in (\ref{xpinn-loss-mse}), thus the training procedure is not completely parallel. The sdPINN makes the training of each sub-domain by the PINN or its improved versions, and thus is in a formation of totally parallel computations and also saves the computational costs. 


$\bullet$ \emph{Effective handling the inverse problem of PDEs}: Since the training in each sub-domain is independent, thus one can choose a sub-domain with high-accuracy, not in the whole domain, to recover the undetermined parameters in PDEs. In Section 4, we will consider the inverse problem of PDEs by the sdPINN explicitly. 
%
%
\subsection{Numerical results}
\subsubsection{The KdV equation revisited}
We revisit the KdV equation (\ref{KdV}) with $b=20$ which is admitted by a translation group $G_{kdv}: x^*=x+2\epsilon, t^*=t+\epsilon, u^*=u$ with the group parameter $\epsilon$. Then for a discrete point $(x_i,t_i,u_i)$ in the initial and boundary conditions, by the transformation group $G_{kdv}$ one has \cite{Dor-2011}
\begin{eqnarray}\label{g-kdv}
&& (x_i,t_i,u_i)\rightarrow(x_i+2\epsilon,t_i+\epsilon,u_i)\rightarrow\cdots\rightarrow(x_i+2k\epsilon,t_i+k\epsilon,u_i),
\end{eqnarray}
where the integer $k$ is the number of group $G_{kdv}$ actions. Moreover, the invariants of $G_{kdv}$ are $I_1:x-2t=C_1, I_2:u=C_2$ with two integrated constants $C_1$ and $C_2$ and the invariant surface condition induced by $G_{kdv}$ is $g:=u_t+2u_x=0$.

\begin{figure}[ht]
	\begin{minipage}{0.5\linewidth}
		\centerline{\includegraphics[width=\textwidth]{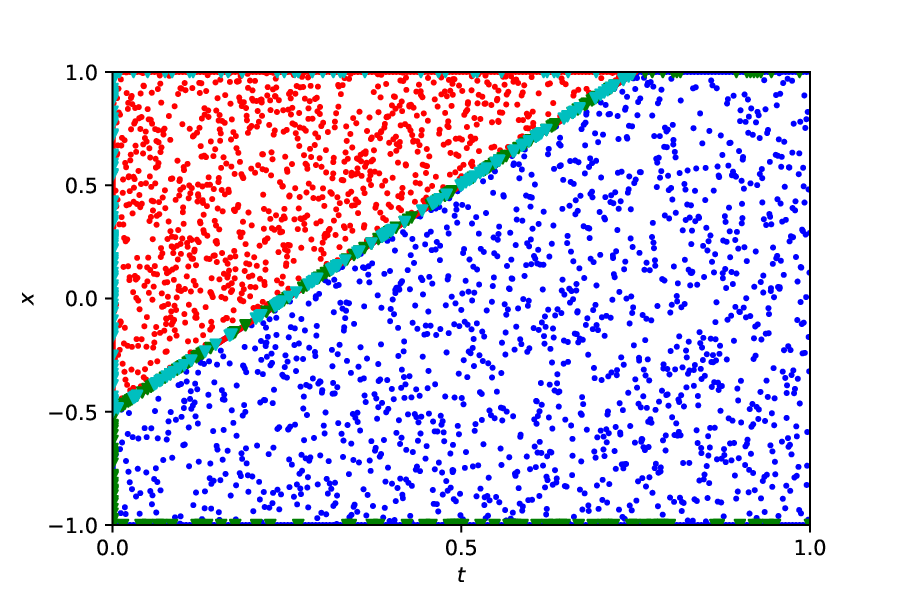}}
        \centerline{(A) sdPINN}
	\end{minipage}
    \begin{minipage}{0.5\linewidth}
		\centerline{\includegraphics[width=\textwidth]{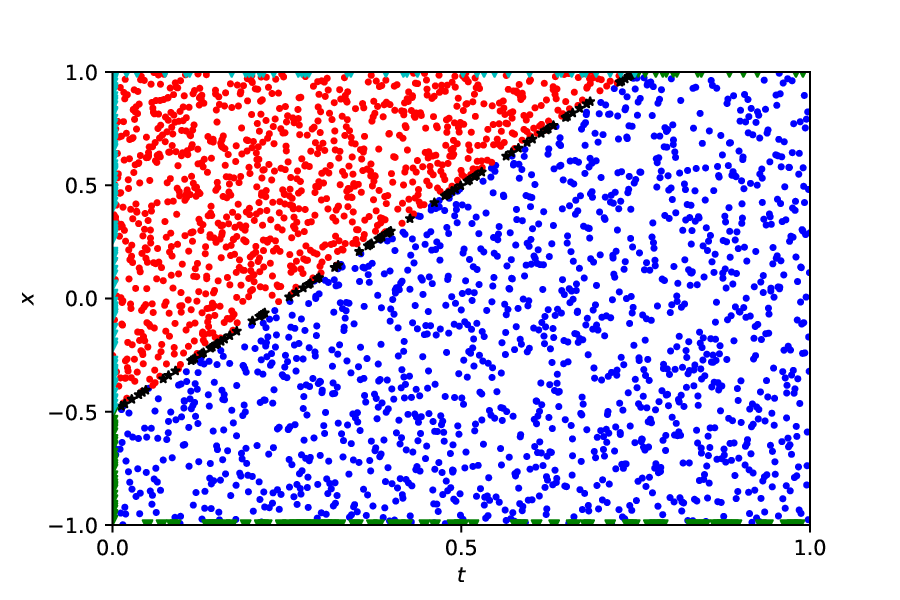}}
        \centerline{(B) XPINN}
	\end{minipage}
	\caption{(Color online) KdV equation: Distributions of training data for the XPINN and sdPINN. The two methods take the same number of training data but do not use the same number of training points due to the random sampling.}
\label{fig4}
\end{figure}

Then following the steps of sdPINN, we choose a discrete point $(-0.5,0,19.7500)$ on the initial condition as the starting point of the invariant $I_1$, then the dividing-line $I_1: x-2t=-0.5$ splits the whole domain $(x,t)\in[-1,1]\times[0,1]$ into two parts which is displayed in Figure \ref{fig4}(A). Furthermore, by the actions (\ref{g-kdv}) induced by $G_{kdv}$, the points on the interface $I_1$ of the two sub-domains are given by $\{(-0.5+2k\epsilon,k\epsilon,19.7500)\}$. Therefore, in the triangular sub-domain in the top left corner, the initial and boundary conditions are all known exactly while for the pentagon sub-domain in the bottom right corner one initial and three boundary conditions are known exactly, thus we perform the sdPINN and sdPINN-isc methods to learn the solutions respectively. Furthermore, we compare the two methods with XPINN whose distributions of training data are displayed in Figure \ref{fig4}(B), where the unique distinction between the two graphs in Figure \ref{fig4} is the interface of two sub-domains where the black interface by XPINN is constrained by the two terms $MSE_{R}$ and $MSE_{u_{avg}}$ in (\ref{xpinn-loss-mse}) while the cyan interface by sdPINN is exactly known.

To compare the PINN, XPINN, sdPINN and sdPINN-isc, we use the same network structure, i.e., $4$ hidden layers with $40$ neurons per layer, random choosing $200$ training points, $2000$ collocation points, the hyperbolic tangent activation function, and deploy the L-BFGS algorithm to minimize the corresponding loss functions and adopt the $L_2$ relative error to measure the effects. Table \ref{tab1} shows the results of five random experiments for each method respectively. The XPINN shows poor performance and the mean $L_2$ relative error arrives at $1.3768$, but the sdPINN gets $1.2526\times10^{-2}$ and the sdPINN-isc gives $6.2293\times10^{-3}$.
\begin{table}[htp]\footnotesize
\captionsetup{width=.9\textwidth,font={footnotesize}}
\caption{The KdV equation: The $L_2$ relative errors by the PINN, XPINN, sdPINN and sdPINN-isc method are performed with five different seeds under the same network structure, 4 hidden layers, and 40 neurons per layer with $N_u=200$ and $N_f=2000$. The $error_1$ and $error_2$ denote the $L_2$ relative errors of the triangular and pentagon sub-domains respectively while $error$ is the $L_2$ relative error of the whole domain.}
\centering
\renewcommand{\arraystretch}{1.2}
\begin{tabular}{c c| ccc c ccc}
\hline
                           &&1&2&3&4&5& Mean \\ \hline
PINN                &$error$&{1.5157e+00}&{1.0504e+00}&{2.8327e-02}&{1.0241e-01}&{9.8929e-01}&{7.3722e-01}\\ \hline
                    &$error$&{1.5195e+00}&{6.5730e-01}&{1.3270e+00}&{1.7388e+00}&{1.6415e+00}&{1.3768e+00}\\
XPINN               &$error_1$&{1.7991e+00}&{1.3575e-01}&{1.2904e+00}&{2.0584e+00}&{1.9432e+00}&{1.4454e+00}\\
                    &$error_2$&{5.6563e-03}&{1.2089e+00}&{1.4140e+00}&{5.9937e-02}&{5.3540e-02}&{5.4840e-01}\\ \hline
                    &$error$&{2.0984e-02}&{1.3100e-02}&{6.7134e-03}&{1.3094e-02}&{8.7391e-03}&{1.2526e-02}\\
sdPINN              &$error_1$&{1.0039e-02}&{1.4776e-02}&{7.0445e-03}&{1.4821e-02}&{5.9641e-03}&{1.0529e-02}\\
                    &$error_2$&{3.5850e-02}&{7.4423e-03}&{5.8082e-03}&{7.1762e-03}&{1.3338e-02}&{1.3923e-02}\\ \hline
                    &$error$&{9.0301e-03}&{7.5230e-03}&{8.9646e-04}&{9.8954e-03}&{3.8015e-03}&{6.2293e-03}\\
sdPINN-isc          &$error_1$&{8.4873e-03}&{8.7747e-03}&{8.5216e-04}&{9.7751e-03}&{4.3717e-03}&{6.4522e-03}\\
                    &$error_2$&{1.0257e-02}&{2.4148e-03}&{9.9820e-04}&{1.0189e-02}&{1.6889e-03}&{5.1095e-03}\\ \hline
\end{tabular}
\label{tab1}
\end{table}

\begin{figure}[htp]
    \begin{minipage}{0.24\linewidth}
		\vspace{3pt}
		\centerline{\includegraphics[width=\textwidth,height=0.8\textwidth]{KDV-5055-uExact.eps}}
        \centerline{A}
	\end{minipage}
	\begin{minipage}{0.24\linewidth}
		\vspace{3pt}
		\centerline{\includegraphics[width=\textwidth,height=0.8\textwidth]{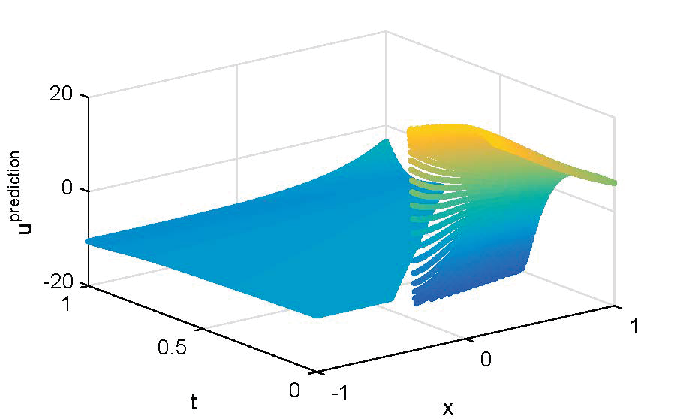}}
        \centerline{B}
	\end{minipage}
	\begin{minipage}{0.24\linewidth}
		\vspace{3pt}
		\centerline{\includegraphics[width=\textwidth,height=0.8\textwidth]{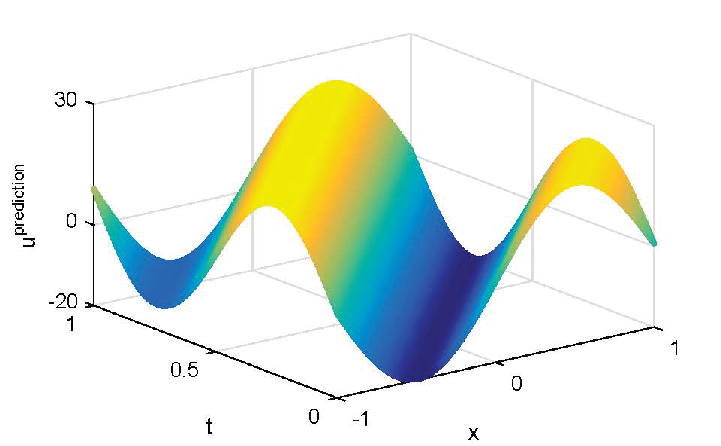}}
        \centerline{C}
	\end{minipage}
    \begin{minipage}{0.24\linewidth}
		\vspace{3pt}
		\centerline{\includegraphics[width=\textwidth,height=0.8\textwidth]{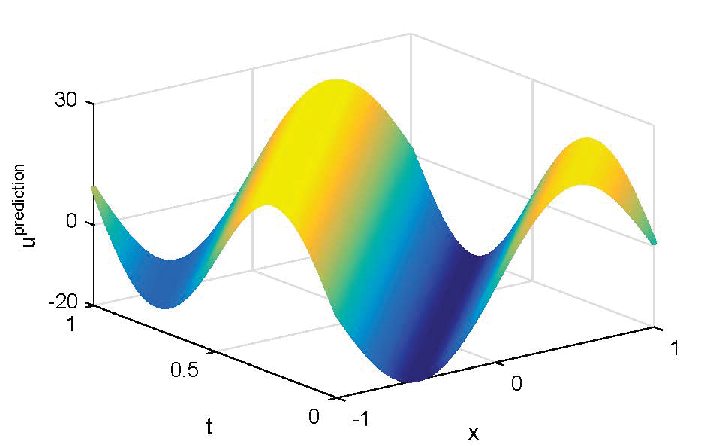}}
        \centerline{D}
	\end{minipage}
    \begin{center}
    \begin{minipage}{0.32\linewidth}
		\vspace{3pt}
		\centerline{\includegraphics[width=\textwidth]{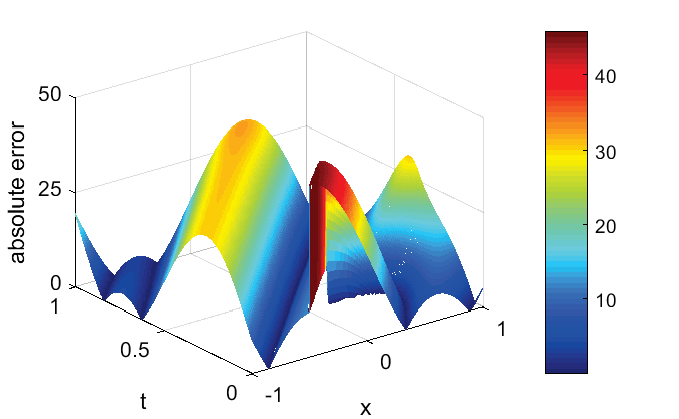}}
        \centerline{E}
	\end{minipage}
    \begin{minipage}{0.32\linewidth}
		\vspace{3pt}
		\centerline{\includegraphics[width=\textwidth]{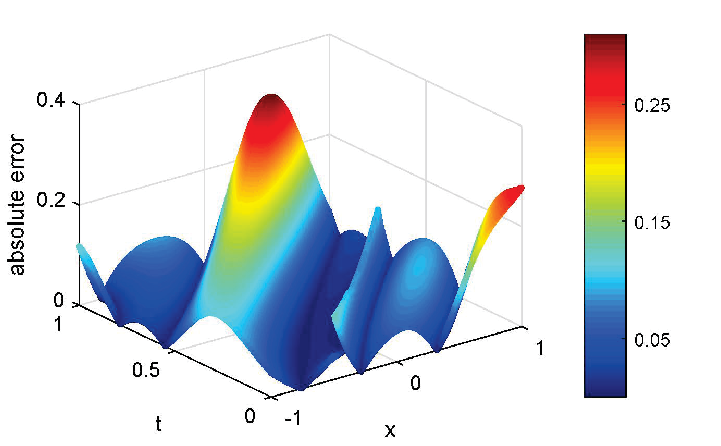}}
        \centerline{F}
	\end{minipage}
    \begin{minipage}{0.32\linewidth}
		\vspace{3pt}
		\centerline{\includegraphics[width=\textwidth]{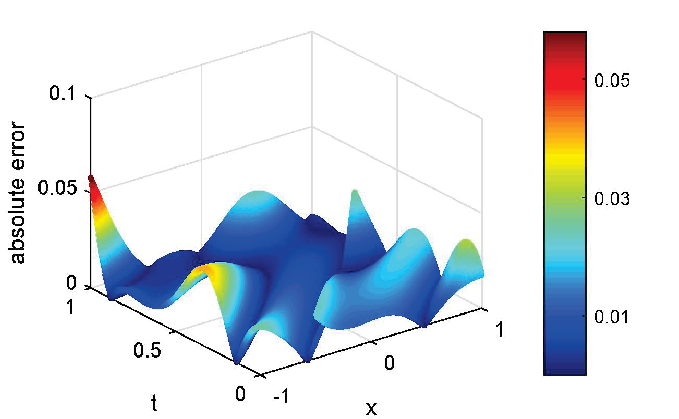}}
        \centerline{G}
	\end{minipage}
    \end{center}
	\caption{(Color online) KdV equation: Comparisons of exact solution and predicted solution and absolute errors of XPINN, sdPINN and sdPINN-isc. (A) Exact solution. (B) Predicted solution by XPINN. (C) Predicted solution by sdPINN. (D) Predicted solution by sdPINN-isc. (E) Absolute error by XPINN. (F) Absolute error by sdPINN. (G) Absolute error by sdPINN-isc.}
\label{fig5}
\end{figure}

Furthermore, we show the graphs of the exact solution, the predicted solutions and the corresponding absolute errors of the XPINN, sdPINN and sdPINN-isc in Figure \ref{fig5}. The predicted solution by XPINN in Figure \ref{fig5}(B) has a big gap with the exact solution in Figure \ref{fig5}(A) where the discontinuous case at the interface emerges, while the predicted solutions by sdPINN and sdPINN-isc in Figure \ref{fig5}(C-D) have no visual differences by eyes with the exact solution. Moreover, the absolute errors in Figure \ref{fig5}(E-G) confirm that the maximal absolute error by XPINN reaches $45.8677$ and the one by sdPINN-isc arrives at $5.8075\times10^{-2}$.

\begin{figure}[htp]
	\begin{minipage}{0.5\linewidth}
		\vspace{3pt}
		\centerline{\includegraphics[width=\textwidth]{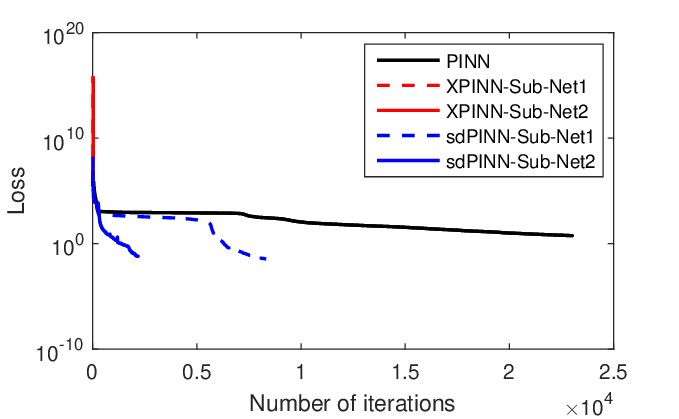}}
        \centerline{A}
	\end{minipage}
	\begin{minipage}{0.5\linewidth}
		\vspace{3pt}
		\centerline{\includegraphics[width=\textwidth]{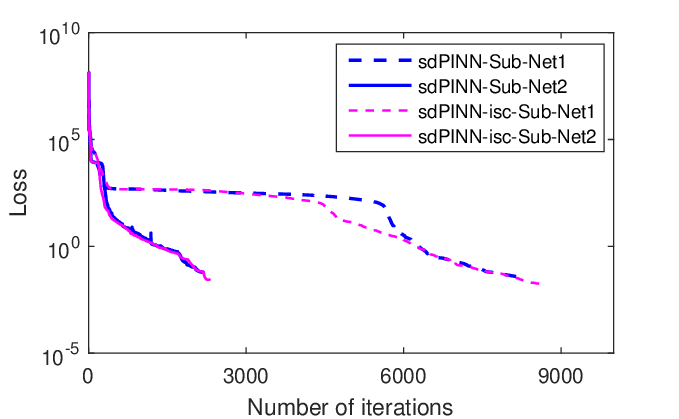}}
        \centerline{B}
	\end{minipage}
	\caption{(Color online) KdV equation: The decrease tendencies of loss functions. (A) Loss reduction of PINN, XPINN and sdPINN. (B) Loss reduction of sdPINN and sdPINN-isc.}
\label{fig6}
\end{figure}

Meanwhile, Figure \ref{fig6}(A) presents the changing tendencies of loss functions of PINN in the whole domain, XPINN and sdPINN in the two sub-domains. The red dotted and solid lines for XPINN always keep big values while the black line for PINN takes a slow downward trend and the final value is greater than one, but the blue lines for sdPINN present an obvious declining trend and the loss values of two sub-domains reach the almost the same order of magnitude. Comparatively, in Figure \ref{fig6}(B), the tendencies of loss function by sdPINN-isc walk almost the same roads as the ones by sdPINN but the pink lines for sdPINN-isc decrease lower than the blue lines of sdPINN where in the triangular sub-domain the sdPINN-isc takes faster decreasing than sdPINN between $4000$ and $6000$ iterations, with final loss values of $1.8248\times10^{-2}$ and $2.7540\times10^{-2}$ in the first and second sub-domains while the final loss values of sdPINN in the two sub-domains are $3.6154\times10^{-2}$ and $6.2554\times10^{-2}$ respectively, which further confirms the good performances of sdPINN-isc in Table \ref{tab1}.

Since the sdPINN has a completely independent node for each sub-domain, then at the end of this subsection, we show that adjusting the neural network architectures of sub-domains can get higher accuracy learned solutions of Eq.(\ref{KdV}). Table \ref{tab2} displays the predicted results of two different neural networks for solving Eq.(\ref{KdV}) where the upper table adopts the completely same situations for the two sub-domains while the lower table employs different neural networks for different sub-domains. Observed that the results in the upper table are the best ones in Table \ref{tab1}, but the $L_2$ relative errors in the lower table still have certain advantages in both two sub-domains, decreasing $30.8865\%$ for the triangular sub-domain and $66.2598\%$ for the pentagon sub-domain respectively.
\begin{table}[htp]\footnotesize
\captionsetup{width=.9\textwidth,font={footnotesize}}
\caption{KdV equation: Two different neural network architectures are used to solve equation. In the upper table, $L_2$ relative error under this network structure is $6.7134\times10^{-3}$. In the lower table, $L_2$ relative error under this network structure is $4.2439\times10^{-3}$.}
\centering
\renewcommand{\arraystretch}{1.2}
\begin{tabular}{ccc c ccc c ccc}
\hline
Sub-domain                      &Triangular&Pentagon\\ \hline
Layers                               &4&4 \\
Neurons                              &40&40 \\
Collocation points                      &2000&2000 \\
Corresponding training points   &200&200 \\
 $L_2$ relative error                      &{7.0445e-03}&{5.8082e-03} \\ \hline
Sub-domain                      &Triangular&Pentagon\\ \hline
Layers                               &4&6 \\
Neurons                              &60&45 \\
Collocation points                      &2000&1500 \\
Corresponding training points   &200&50 \\
 $L_2$ relative error                      &{4.8687e-03}&{1.9597e-03} \\
\hline
\end{tabular}
\label{tab2}
\end{table}

\subsubsection{Non-linear viscous fluid equation}
The second example is the non-linear viscous fluid equation \cite{sk-1974,jan}
\begin{eqnarray} \label{fluid}
&& u_t+uu_{x}-(u^3u_{x})_{x}=\mu(t,x),~~~~(x,t)\in[-1,1]\times[0.5,1],
\end{eqnarray}
which has an exact solution $u_{nvf}=t\big(20\,\text{sech}(x/t^2)+(x/t^2)^2\big)$ and then $\mu(x,t)$ as well as the initial and the boundary conditions are forced by the solution $u_{nvf}$. Moreover, Eq.(\ref{fluid}) admits a scaling group $G_{nvf}: x^*=\epsilon^2x, t^*=\epsilon t, u^*=\epsilon u$, where $\epsilon$ is the group parameter. Then for a discrete point $(x_i,t_i,u_i)$ on the initial and boundary conditions, by the transformation group $G_{nvf}$ one obtains \cite{Dor-2011}
\begin{eqnarray}
&&\no (x_i,t_i,u_i)\rightarrow(\epsilon^2x_i,\epsilon t_i,\epsilon u_i)\rightarrow\cdots\rightarrow(\epsilon^{2k}x_i,\epsilon^kt_i,\epsilon^ku_i),
\end{eqnarray}
where the integer $k$ is the number of group $G_{nvf}$ actions. Furthermore, the two invariants of $G_{nvf}$ are $I_1:=x/t^2=C_1, I_2:=u/t=C_2$ and the invariant surface condition of $G_{nvf}$ is $g_{nvf}:=tu_t+2xu_x-u=0$.

\begin{figure}[htp]
\begin{minipage}{0.33\linewidth}
		\centerline{\includegraphics[width=\textwidth,height=0.8\textwidth]{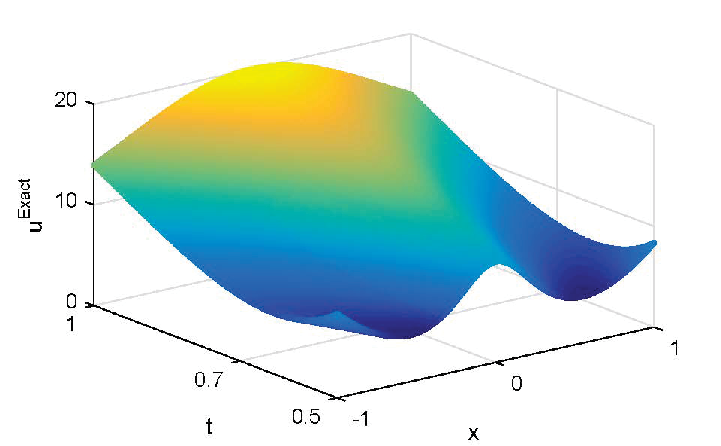}}
        \centerline{(A)}
	\end{minipage}
    \begin{minipage}{0.33\linewidth}
		\centerline{\includegraphics[width=\textwidth,height=0.8\textwidth]{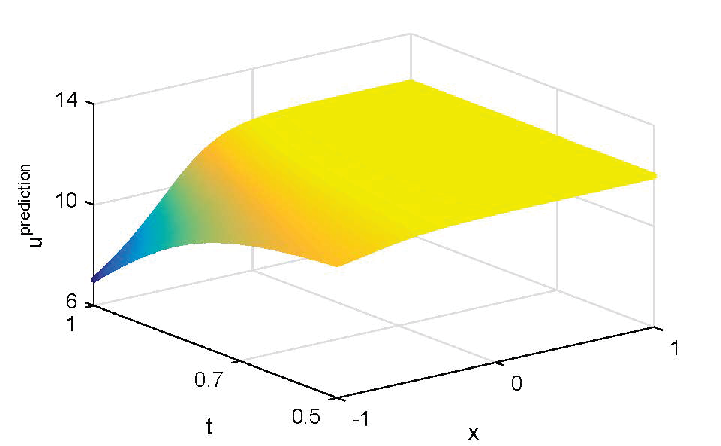}}
        \centerline{(B)}
    \end{minipage}
    \begin{minipage}{0.33\linewidth}
		\centerline{\includegraphics[width=\textwidth,height=0.8\textwidth]{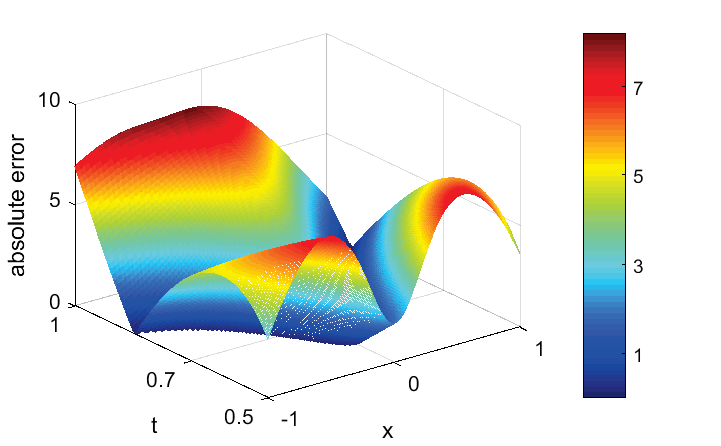}}
        \centerline{(C)}
	\end{minipage}
	\caption{(Color online) Non-linear viscous fluid equation:  (A) Exact solution. (B) Predicted solution by PINN. (C) Absolute errors by PINN.}
\label{fig7}
\end{figure}

We first use the PINN with 4 hidden layers with 40 neurons per layer to learn the solution $u_{nvf}$ where the spatial domain $x\in[-1,1]$ and the temporal domain $t\in[0.5,1]$ are discretized into $N_x=400$ and $N_t=100$ discrete equidistant points respectively, and the 1000 collocation points via the Latin hypercube sampling \cite{ms-1987} and the randomly choosing 100 training points on the initial and boundary conditions. Observed that the PINN does a poor job in the training where the predicted solution in Figure \ref{fig7}(B) is distorted with the exact solution  in Figure \ref{fig7}(A) and the maximal absolute error in Figure \ref{fig7}(C) arrives at $8.1902$.

\begin{figure}[htp]
    \begin{minipage}{0.5\linewidth}
		\centerline{\includegraphics[width=\textwidth]{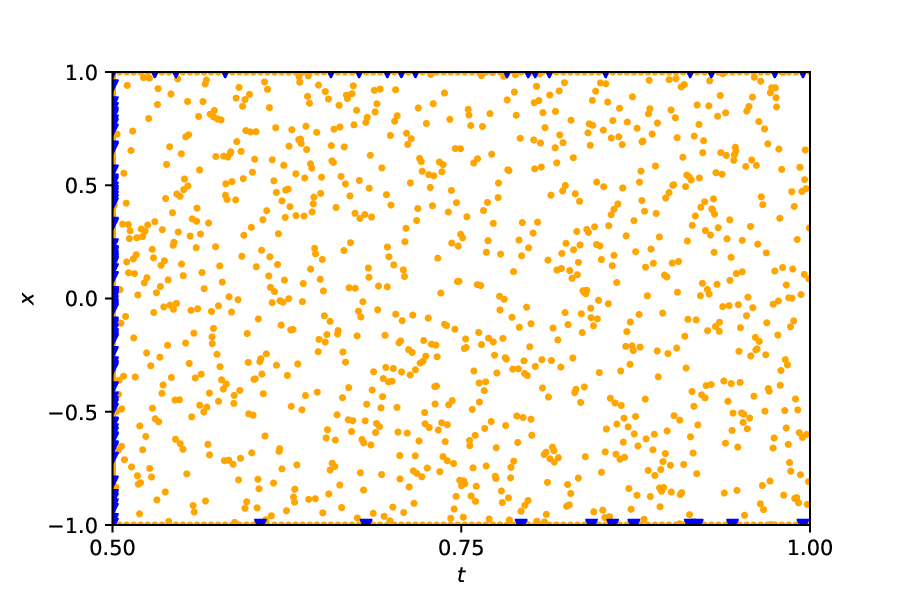}}
        \centerline{(A)}
	\end{minipage}
	\begin{minipage}{0.5\linewidth}
		\centerline{\includegraphics[width=\textwidth]{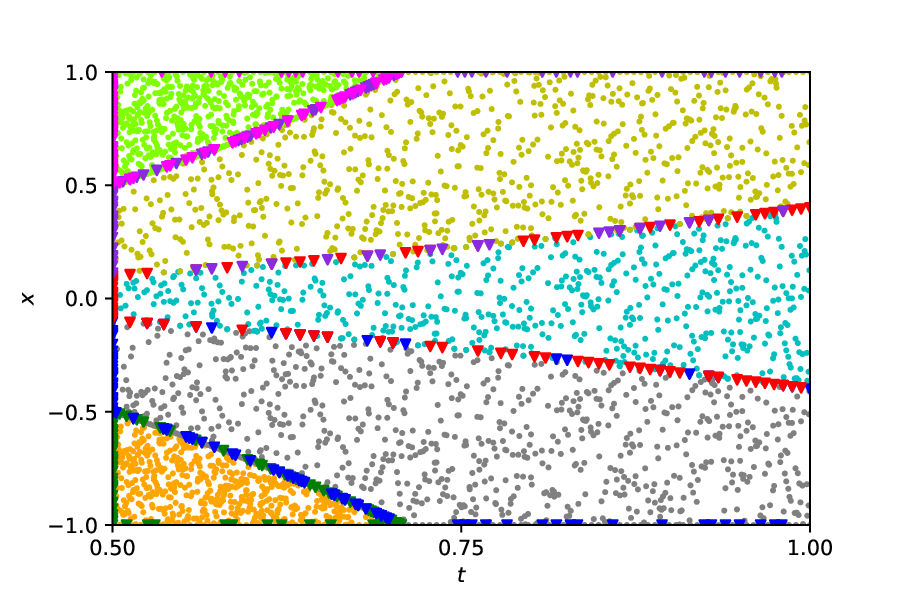}}
        \centerline{(B) sdPINN}
	\end{minipage}
	\caption{(Color online) Non-linear viscous fluid equation: Domain partition and distributions of training data for the XPINN and sdPINN.} 
\label{fig8}
\end{figure}

Next, we deploy the sdPINN to learn the solution $u_{nvf}$ of Eq.(\ref{fluid}). Here we choose four discrete points $(0.5,0.5,4.6580)$, $(0.5,0.1,9.3301)$, $(0.5,-0.1,9.3301)$ and $(0.5,-0.5,4.6580)$ on the initial and boundary conditions as the starting points to determine four dividing-lines via the invariant $I_1:=x/t^2=C_1$, i.e. $t=-\sqrt{0.5x}$, $t=-\sqrt{2.5x}$, $t=\sqrt{2.5x}$ and $t=\sqrt{0.5x}$.  Then the entire domain in Figure \ref{fig8}(A), where the PINN works, is divided into five sub-domains shown in Figure \ref{fig8}(B).
More importantly, the points on the four interfaces of sub-domains are labeled by means of the four starting points and group $G_{nvf}$, i.e.
\begin{eqnarray}
&&\no t=-\sqrt{0.5x}: \{(0.5\epsilon^{2k},0.5\epsilon^k,4.6580\epsilon^k)\},~~~ t=-\sqrt{2.5x}: \{(0.5\epsilon^{2k},0.1\epsilon^k,9.3301\epsilon^k)\},\\
&&\no t=\sqrt{2.5x}: \{(0.5\epsilon^{2k},-0.1\epsilon^k,9.3301\epsilon^k)\},~~~ t=\sqrt{0.5x}: \{(0.5\epsilon^{2k},-0.5\epsilon^k,4.6580\epsilon^k)\},
\end{eqnarray}
where the integer $k$ is the number of group $G_{nvf}$ actions. Comparatively, the interfaces by XPINN with the same sub-domains in Figure \ref{fig8}(B) are unknown and thus constrained by two terms $MSE_{R}$ and $MSE_{u_{avg}}$ in (\ref{xpinn-loss-mse}).

To compare the XPINN, sdPINN and sdPINN-isc methods, we perform the training under the same conditions, i.e., 1000 collocation points and 100 training points, a hyperbolic tangent activation function, and the neural networks with 4 hidden layers of 20 neurons per layer. Table \ref{tab3} shows the results of the PINN, XPINN, sdPINN and sdPINN-ics with five random experiments, where the mean $L_2$ relative errors of sdPINN and sdPINN-isc in the whole domain reach $3.3663\times10^{-3}$ and $3.1520\times10^{-4}$ accuracies respectively, while XPINN only arrives at $3.6165\times10^{-1}$ and keeps the same order of magnitude as the PINN.
\begin{table}[htp]\footnotesize
\captionsetup{width=0.9\textwidth,font={footnotesize}}
\caption{Non-linear viscous fluid equation: XPINN, sdPINN and adding invariant surface condition to sdPINN (sdPINN-isc) are averaged for $L_2$ relative errors of five different seeds under the same network structure, 4 hidden layers, and $20$ neurons per layer with $N_u=100$ and $N_f=1000$.}
\centering
\renewcommand{\arraystretch}{1.2}
\begin{tabular}{c c| ccc c ccc}
\hline
                &&1&2&3&4&5&Mean \\ \hline
PINN       &$error$&{3.5069e-01}&{3.2503e-01}&{6.0695e-01}&{5.7266e-01}&{6.6228e-01}&{5.0352e-01}\\\hline
           &$error$&{3.8467e-01}&{3.8722e-01}&{3.0677e-01}&{3.3259e-01}&{3.9699e-01}&{3.6165e-01}\\
           &$error_1$&{5.8623e-01}&{4.5810e-01}&{5.2931e-01}&{5.7969e-01}&{3.5895e-01}&{5.0246e-01}\\
           &$error_2$&{4.1218e-01}&{5.0984e-01}&{3.1914e-01}&{4.6787e-01}&{3.7285e-01}&{4.1638e-01}\\
XPINN      &$error_3$&{1.2999e-01}&{2.4000e-01}&{1.4182e-01}&{1.7046e-01}&{1.5458e-01}&{1.6737e-01}\\
           &$error_4$&{5.5222e-01}&{3.1008e-01}&{4.2970e-01}&{3.2799e-01}&{6.0434e-01}&{4.4487e-01}\\
           &$error_5$&{1.2538e-01}&{1.3178e+00}&{7.6299e-02}&{2.5310e-01}&{3.1734e-01}&{4.1798e-01}\\ \hline
           &$error$&{2.7477e-03}&{5.2171e-03}&{3.1270e-03}&{2.0999e-03}&{3.6397e-03}&{3.3663e-03}\\
           &$error_1$&{7.5885e-04}&{8.2275e-05}&{3.4857e-04}&{5.2011e-04}&{5.1288e-04}&{4.4453e-04}\\
           &$error_2$&{1.0798e-03}&{6.5323e-04}&{7.6023e-04}&{2.5712e-03}&{2.8386e-03}&{1.5806e-03}\\
sdPINN     &$error_3$&{2.7851e-03}&{8.1222e-03}&{3.4591e-03}&{1.9918e-03}&{4.0553e-03}&{4.0827e-03}\\
           &$error_4$&{3.7763e-03}&{1.6305e-03}&{4.0974e-03}&{1.7737e-03}&{3.8911e-03}&{3.0338e-03}\\
           &$error_5$&{6.7429e-04}&{6.0788e-04}&{3.2608e-04}&{1.0506e-03}&{4.0705e-04}&{6.1319e-04}\\ \hline
           &$error$&{2.0538e-04}&{3.4933e-04}&{4.6283e-04}&{2.2738e-04}&{3.3105e-04}&{3.1520e-04}\\
           &$error_1$&{2.2811e-05}&{6.4759e-05}&{2.1777e-05}&{5.8473e-05}&{8.3448e-05}&{5.0253e-05}\\
           &$error_2$&{2.4017e-04}&{4.5667e-04}&{6.9015e-04}&{1.0007e-04}&{3.8982e-04}&{3.7538e-04}\\
sdPINN-isc &$error_3$&{1.6754e-04}&{3.2496e-04}&{4.3258e-04}&{3.3343e-04}&{2.8470e-04}&{3.0864e-04}\\
           &$error_4$&{2.2331e-04}&{2.6212e-04}&{8.6886e-05}&{1.2049e-04}&{3.3673e-04}&{2.0591e-04}\\
           &$error_5$&{2.5514e-05}&{2.5958e-05}&{1.3334e-04}&{1.3647e-04}&{2.4629e-04}&{1.1351e-04}\\ \hline
\end{tabular}
\label{tab3}
\end{table}

\begin{figure}[htp]
    \begin{center}
    \begin{minipage}{0.24\linewidth}
		\vspace{3pt}
        \centerline{\includegraphics[width=\textwidth,height=0.8\textwidth]{NF-Exact.eps}}
        \centerline{A}
	\end{minipage}
	\begin{minipage}{0.24\linewidth}
		\vspace{3pt}
		\centerline{\includegraphics[width=\textwidth,height=0.8\textwidth]{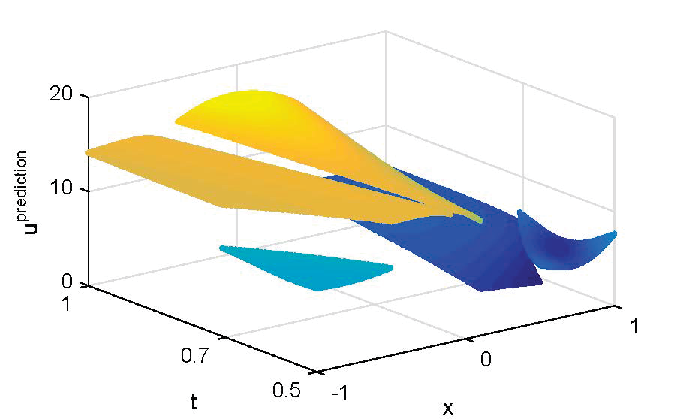}}
        \centerline{B}
	\end{minipage}
	\begin{minipage}{0.24\linewidth}
		\vspace{3pt}
		\centerline{\includegraphics[width=\textwidth,height=0.8\textwidth]{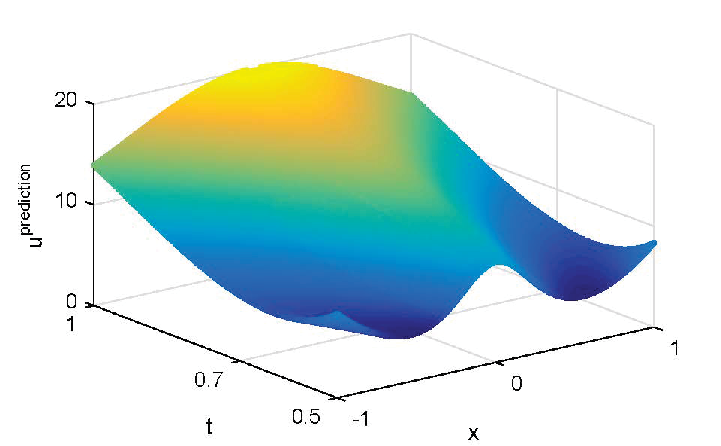}}
        \centerline{C}
	\end{minipage}
    \begin{minipage}{0.24\linewidth}
		\vspace{3pt}
		\centerline{\includegraphics[width=\textwidth,height=0.8\textwidth]{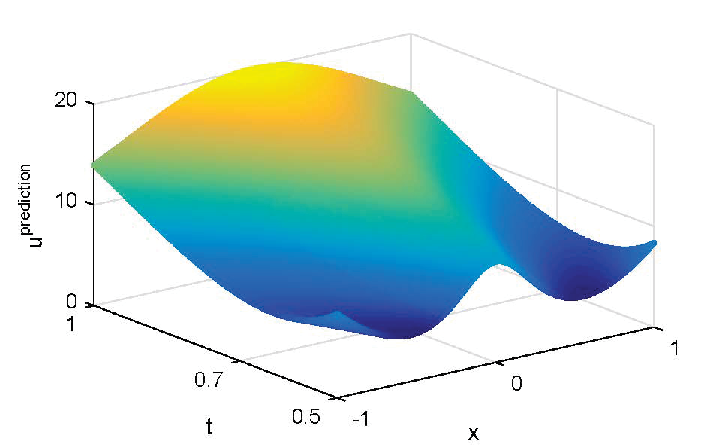}}
        \centerline{D}
	\end{minipage}
    \end{center}
    \begin{center}
    \begin{minipage}{0.32\linewidth}
		\vspace{3pt}
		\centerline{\includegraphics[width=\textwidth]{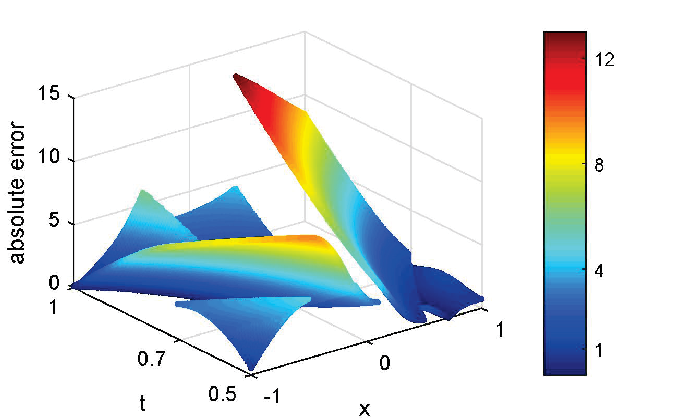}}
        \centerline{E}
	\end{minipage}
    \begin{minipage}{0.32\linewidth}
		\vspace{3pt}
		\centerline{\includegraphics[width=\textwidth]{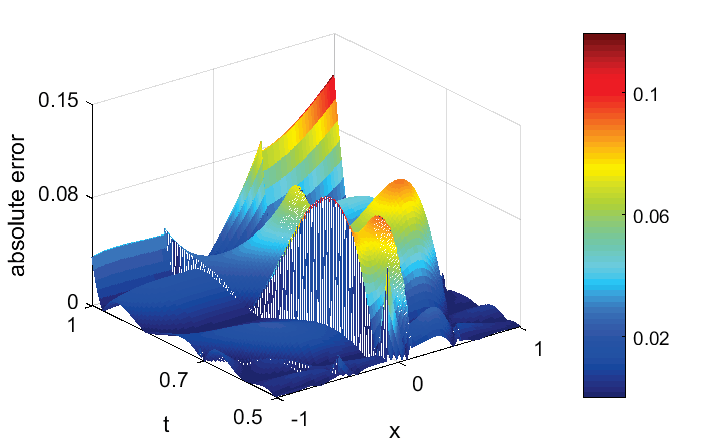}}
        \centerline{F}
	\end{minipage}
    \begin{minipage}{0.32\linewidth}
		\vspace{3pt}
		\centerline{\includegraphics[width=\textwidth]{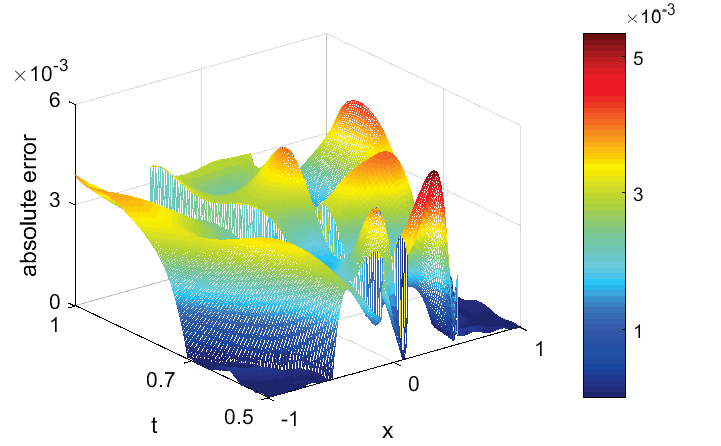}}
        \centerline{G}
	\end{minipage}
    \end{center}
	\caption{(Color online) Non-linear viscous fluid equation: comparison of exact solution and predicted solution and absolute errors of PINN, sdPINN and sdPINN-isc. (A) Exact solution. (B) Predicted solution by XPINN. (C) Predicted solution by sdPINN. (D) Predicted solution by sdPINN-isc. (E) Absolute error by XPINN. (F) Absolute error by sdPINN.(G) Absolute error by sdPINN-isc.}
\label{fig9}
\end{figure}

Furthermore, Figure \ref{fig9} shows the exact solution, the predicted solutions for XPINN, sdPINN and sdPINN-isc as well as their absolute errors. The predicted solution by XPINN in Figure \ref{fig9}(B) has an obvious gap with the exact solution in Figure \ref{fig9}(A) and the maximal absolute error in Figure \ref{fig9}(E) reaches $13.0066$. However, the performances of sdPINN and sdPINN-isc in Figure \ref{fig9}(C-D) are comparatively better than XPINN and PINN, in particular, the absolute errors of the predicted solution by sdPINN-ics in Figure \ref{fig9}(F) arrives at $10^{-3}$ order of magnitude.

\begin{figure}[htp]
    \begin{minipage}{0.33\linewidth}
		\vspace{3pt}
		\centerline{\includegraphics[width=\textwidth]{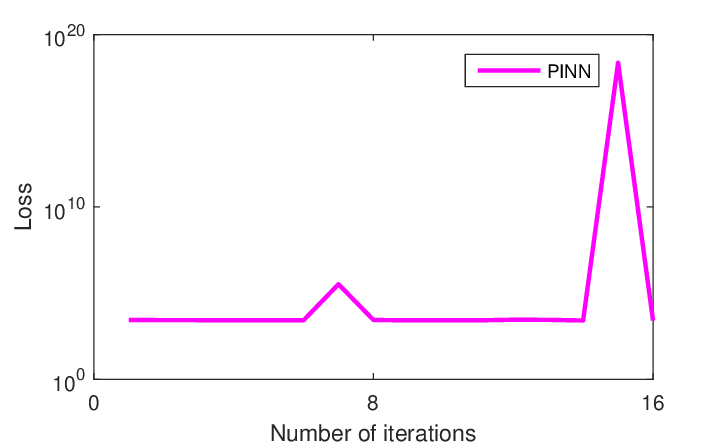}}
        \centerline{A}
	\end{minipage}
	\begin{minipage}{0.33\linewidth}
		\vspace{3pt}
		\centerline{\includegraphics[width=\textwidth]{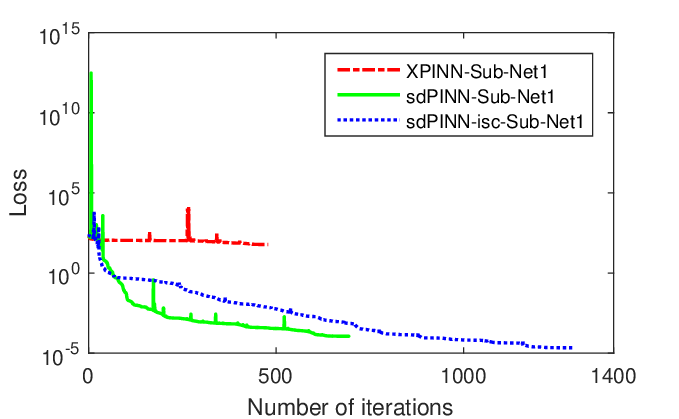}}
        \centerline{B}
	\end{minipage}
	\begin{minipage}{0.33\linewidth}
		\vspace{3pt}
		\centerline{\includegraphics[width=\textwidth]{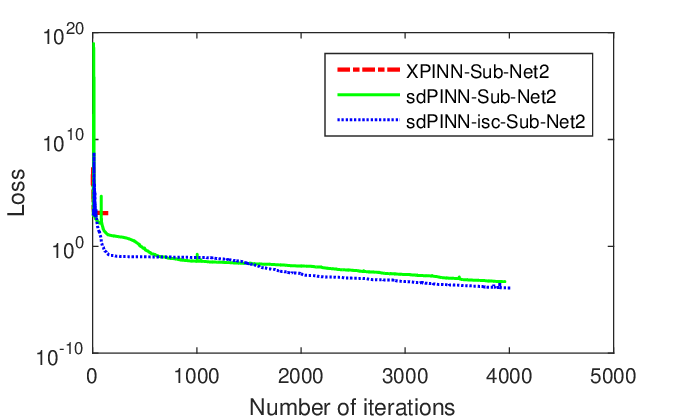}}
        \centerline{C}
	\end{minipage}
    \begin{minipage}{0.33\linewidth}
		\vspace{3pt}
		\centerline{\includegraphics[width=\textwidth]{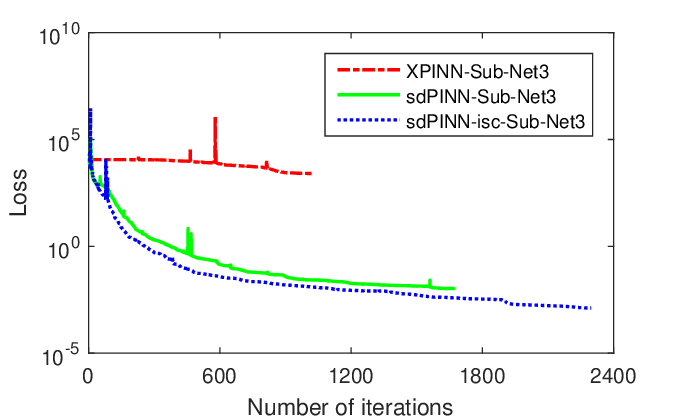}}
        \centerline{D}
	\end{minipage}
    \begin{minipage}{0.33\linewidth}
		\vspace{3pt}
		\centerline{\includegraphics[width=\textwidth]{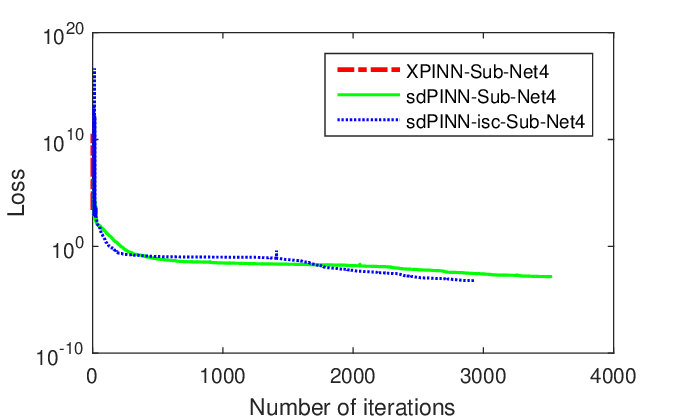}}
        \centerline{E}
	\end{minipage}
    \begin{minipage}{0.33\linewidth}
		\vspace{3pt}
		\centerline{\includegraphics[width=\textwidth]{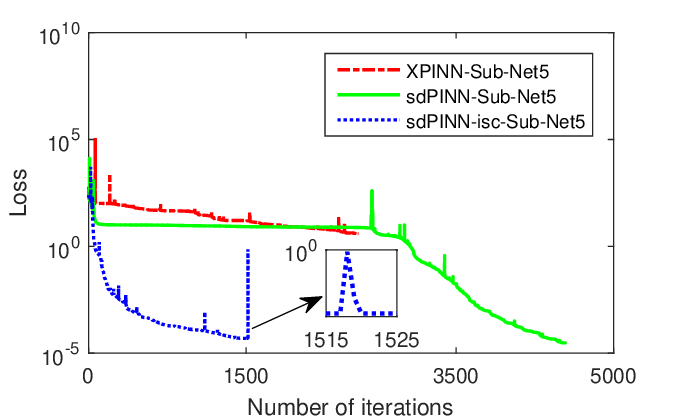}}
        \centerline{F}
	\end{minipage}
	\caption{(Color online) Non-linear viscous fluid equation: The decrease of loss of equation. (A) Loss reduction of PINN. (B-F) Denote the loss drop of the three methods XPINN, sdPINN and sdPINN-isc in five sub-domains respectively.}
\label{fig10}
\end{figure}

Meanwhile, the performances of loss functions of the four methods in Figure \ref{fig10} also confirm the good jobs of sdPINN and sdPINN-isc. The loss function of PINN in Figure \ref{fig10}(A) expresses a divergence tendency while the ones of XPINN in the five sub-domains in Figure \ref{fig10}(B-F) display a slowly decreasing status and finally terminate at a big value. The sdPINN and sdPINN-ics walk a similar way but the sdPINN-isc decreases a bit faster than sdPINN, in particular, in the fifth sub-domain, sdPINN-isc stops at $4.9280\times10^{-5}$ with $1544$ iterations but sdPINN completes the training at $3.0341\times10^{-5}$ with $4541$ iterations. 

In Table \ref{tab4}, we show the flexibility of sdPINN where two different neural network architectures are used to solve Eq.(\ref{fluid}). In the upper table, the partitioned five sub-domains adopt completely the same situations while in the lower table, different sub-domains take different neural networks, and different number of training data to obtain high-accuracy solutions. It is observed that the total $L_2$ relative errors in the lower table have one order of magnitude improvement than the upper ones which are the best ones in Table \ref{tab3}, from $2.7477\times10^{-3}$ to $5.7938\times10^{-4}$.
 \begin{table}[htp]\footnotesize
\captionsetup{width=.9\textwidth,font={footnotesize}}
\caption{Non-linear viscous fluid equation: Two different neural network architectures used to solve equation. In the upper table, $L_2$ relative error under this network structure is $2.7477\times10^{-3}$. In the lower table, $L_2$ relative error under this network structure is $5.7938\times10^{-4}$.}
\centering
\renewcommand{\arraystretch}{1.2}
\begin{tabular}{ccc c ccc c ccc}
\hline
Sub-domain                   &1&2&3&4&5 \\ \hline
Layers                          &4&4&4&4&4 \\
Neurons                              &20&20&20&20&20 \\
Collocation points                      &1000&1000&1000&1000&1000 \\
Corresponding training points   &100&100&100&100&100 \\
Rel. $L_2$ error                      &{7.5885e-04}&{1.0798e-03}&{2.7851e-03}&{3.7763e-03}&{6.7429e-04} \\ \hline
Sub-domain                       &1&2&3&4&5 \\ \hline
Layers                               &6&2&4&2&4 \\
Neurons                              &10&20&40&40&20 \\
Collocation points                      &3000&1000&1000&1500&2000 \\
Corresponding training points   &150&200&50&150&200 \\
Rel. $L_2$ error                      &{4.1924e-04}&{7.3947e-04}&{4.9525e-04}&{5.1527e-04}&{3.6983e-04} \\
\hline
\end{tabular}
\label{tab4}
\end{table}

\section{Inverse problem}
Utilizing a neural network for the inverse problem of PDEs means to discover the undetermined parameters in PDEs as well as their numerical solution. In the traditional methods for the inverse problem, a proper number of labeled data in the interior domain of PDEs is required and also takes high computational costs. In addition, the training usually is performed in the whole domain of PDEs and thus needs more training points. Therefore, by considering the above two issues for the PDEs admitting a Lie symmetry group, we leverage the idea of the sdPINN method to introduce a new method where the labeled data in the interior domain are generated by acting the symmetry group on the data of the initial and boundary conditions and the training is only done in a sub-domain, not the whole domain of PDEs.
\subsection{sdPINN for the inverse problem}
We take the following $r$-th order PDE
\begin{eqnarray} \label{eqn1-in}
&& f: =u_t+\mathcal {N}[u;\lambda]=0,~~~~~t\in [0, T],~~x\in [a,b],
\end{eqnarray}
and the initial and boundary conditions (\ref{ib}) as an example to introduce the main idea of sdPINN for the inverse problem, where $\mathcal {N}[u;\lambda]$ is a smooth function of $u$ and its $x$-derivatives up to $r$th order where a parameter $\lambda$ is to be determined.

Suppose that Eq.(\ref{eqn1-in}) is admitted by the symmetry group (\ref{group}), then following the idea of sdPINN for the forward problem, we have found the dividing-lines which divide the whole domain into several non-overlapping sub-domains whose the interfaces of any two neighbouring sub-domains are exactly known. Thus, we choose one sub-domain $\Omega_p$ as the target training domain for the inverse problem.
Specifically, let $(x_{ib},t_{ib},u(x_{ib},t_{ib}))$ be a discrete point on the initial and boundary conditions of $\Omega_p$. Then, by the property that a symmetry group can map a solution to another solution of the same equation, we obtain a labeled data set $\mathcal {S}_{ld}=\{(x^{(i)},t^{(i)},u^{(i)})\}_{i=1}^{N_{p}}$ in the interior domain of $\Omega_p$, where
\begin{eqnarray}
&&\no x^{(i)}=x_{ib}+\varepsilon\,i\,\xi+O(\varepsilon^2),t^{(i)}=t_{ib}+\varepsilon\,i\,\tau +O(\varepsilon^2),u^{(i)}=u(x_{ib},t_{ib})+\varepsilon\,i\,\eta+O(\varepsilon^2),
\end{eqnarray}
where $(x^{(0)},t^{(0)},u^{(0)})=(x_{ib},t_{ib},u(x_{ib},t_{ib}))$.

Denoted by $\mathcal {S}_{ib}$ the set of $N_{ib}$ chosen training data on the initial and boundary conditions, then the total labeled data for the inverse problem are collected in $\mathcal {S}_{total}=\mathcal {S}_{ib}\cup\mathcal {S}_{ld}$. Following the framework of neural networks for the forward problem, the loss function for the inverse problem becomes
\begin{eqnarray} \label{invaloss-inv}
&&MSE^{inv}=w_p MSE_p+w_{u_p} MSE_{u_p}+w_{f_p} MSE_{f_p},
\end{eqnarray}
where $MSE_{u_p}$ and $MSE_{f_p}$ are given in (\ref{sdPINN}), and the additional term $MSE_p$ defined by
\begin{eqnarray}
&&\no MSE_p=\frac{1}{N_{p}}\sum_{i=1}^{N_p}\,|\widetilde{u}(x^{(i)},t^{(i)})-u^{(i)}|^2,
\end{eqnarray}
measures the errors between the exact solution and the learned solution in $\mathcal {S}_{ld}$. Then, one deploys the usual optimization algorithms to search for the minimal value of $MSE^{inv}$ to give the undetermined parameters as well as the numerical solutions.

When comparing with the vanilla PINN method, the proposed method for data-discovery of PDEs has two distinctive merits. One is that the training is performed only in a sub-domain whose initial and boundary conditions are exactly known by means of the symmetry group of Eq.(\ref{eqn1-in}), while the PINN works in the whole domain and thus needs more training points. The other is the set of labeled data in the interior sub-domain which are obtained by acting the Lie symmetry group  (\ref{group}) on the labeled data $\mathcal {S}_{ib}$ of the initial and boundary conditions, while the PINN needs extra labeled data from external supports.

\subsection{Numerical results}
\begin{figure}[htp]
	\begin{minipage}{0.34\linewidth}
		\vspace{3pt}
		\centerline{\includegraphics[width=\textwidth]{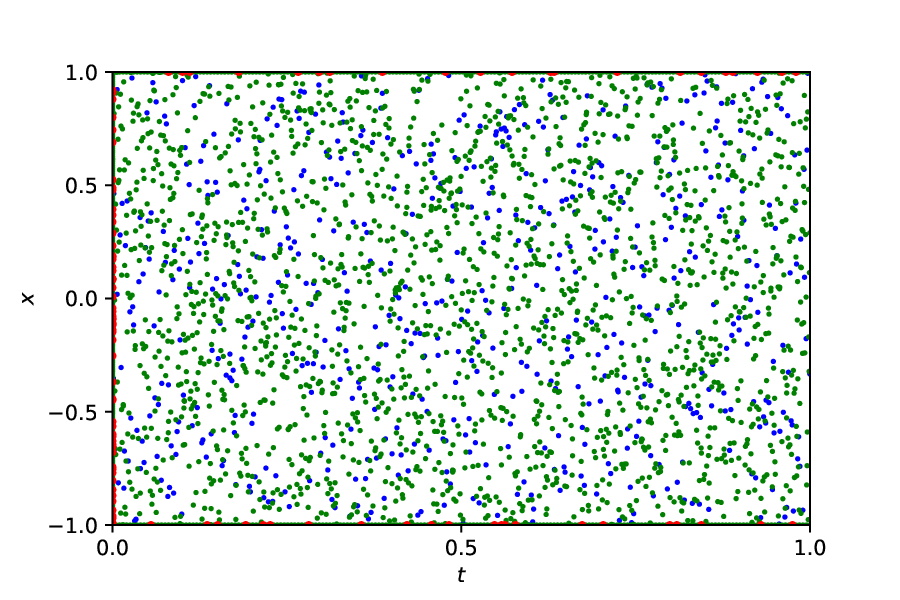}}
        \vspace{5pt}
        \centerline{(A) PINN}
	\end{minipage}
    \begin{minipage}{0.34\linewidth}
		\vspace{3pt}
		\centerline{\includegraphics[width=\textwidth]{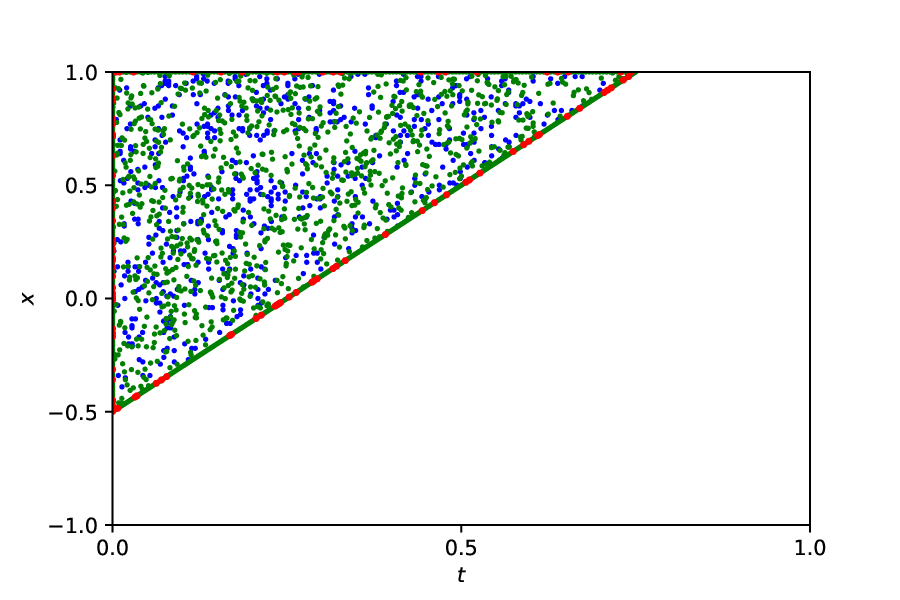}}
        \vspace{5pt}
        \centerline{(B) sdPINN}
	\end{minipage}
    \hspace{0.04\linewidth}
    \begin{minipage}{0.23\linewidth}
		\vspace{3pt}
		\centerline{\includegraphics[width=1.3\textwidth, height=0.86\textwidth]{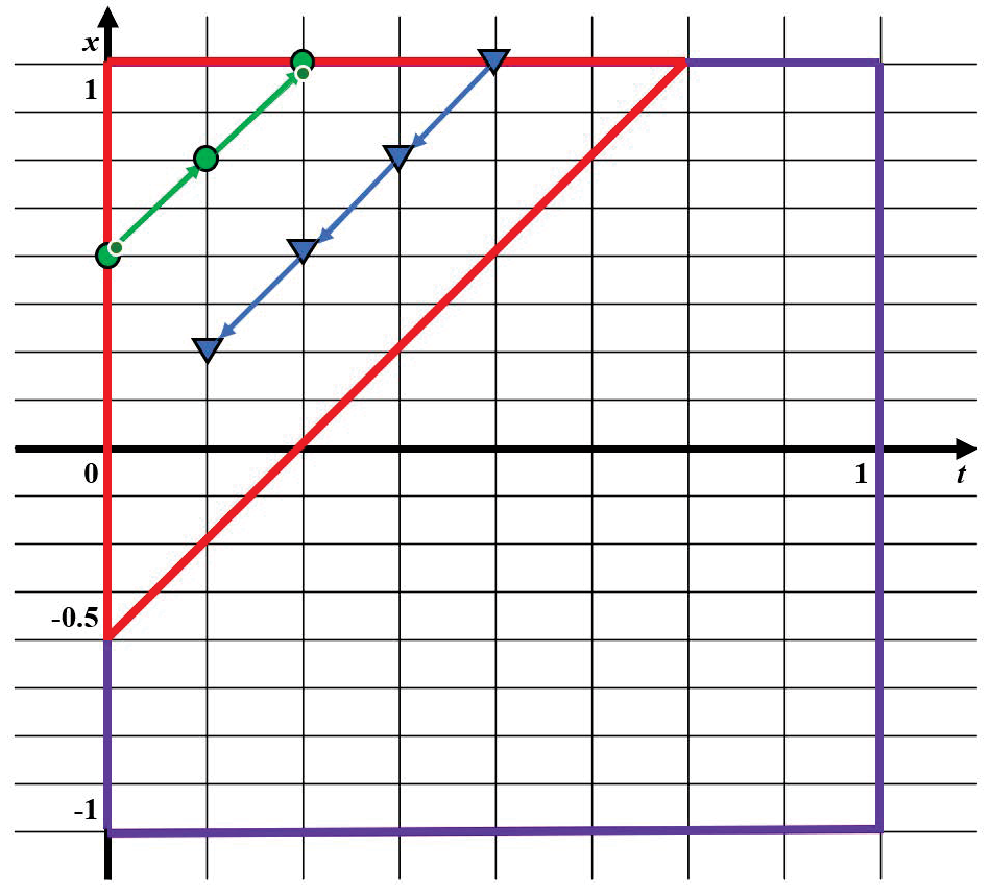}}
        \vspace{12.5pt}
        \centerline{(C) Generating path}
	\end{minipage}
	\caption{(Color online) Case of $b=5$: (A) Distribution of training data for the PINN. (B) Distribution of training data for the sdPINN. (C) An explicit process of generating interior labeled data from the data on initial and boundary conditions.}
\label{fig11}
\end{figure}

We consider the KdV equation (\ref{KdV}) by adding a parameter artificially, i.e.
\begin{eqnarray} \label{KdV-p}
&& u_t+\lambda uu_{x}+u_{xxx}=\mu(x,t),~~~~(x,t)\in[-1,1]\times[0,1],
\end{eqnarray}
where $\lambda$ is an undetermined constant. We assume that Eq.(\ref{KdV-p}) with $\lambda=1$ has a solution $u_{kdv}=\left(x-2t\right)^2+b\sin(\pi(x-2t))$ and $\mu(x,t)$ is enforced by the solution $u_{kdv}$. Moreover, as shown in Figure \ref{fig11}(B), we adopt the same domain decomposition as the forward problem of Eq.(\ref{KdV}) and choose the top left corner sub-domain in Figure \ref{fig11}(B) as the training domain, while the PINN works in the whole domain in Figure \ref{fig11}(A).

In order to obtain the training data, we first discretize the spatial domain $x\in[-1,1]$ and the time domain $t\in[0,1]$ into $N_x=400$ and $N_t=200$ discrete equidistance points respectively, and then obtain a data set $\mathcal {S}_{ib}$ of $800$ points. Then with the translation group $G_{kdv}: x^*=x+2\epsilon, t^*=t+\epsilon, u^*=u$, from the data in $\mathcal {S}_{ib}$, one gets the interior labeled data set $\mathcal {S}_{ld}$ via the following two pathes,
\begin{eqnarray}
&&\no \bullet: (x_i,0,u_i)\rightarrow(x_i+2\epsilon,\epsilon,u_i)\rightarrow\cdots\rightarrow(x_i+2k\epsilon,k\epsilon,u_i); \\
&&\no \blacktriangledown: (1,t_j,u_i)\rightarrow(1-2\epsilon,t_j-\epsilon,u_i)\rightarrow\cdots\rightarrow(1-2k\epsilon,t_j-k\epsilon,u_i),
\end{eqnarray}
where $k$ is a positive integer.
More specifically, Figure \ref{fig11}(C) describes the generating procedure from the data in $\mathcal {S}_{ib}$ by the group $G_{kdv}$, where the red line denotes the generating path starting from a labeled point on the initial condition at $t=0$ and the blue line stands for the case of the boundary condition at $x=1$. Finally, we use the L-BFGS algorithm to optimize the parameters of network to minimize the loss function (\ref{invaloss-inv}). In what follows, we consider two cases of $b=5$ and $b=20$ in solution $u_{kdv}$.
\subsubsection{Case of $b=5$}
We calibrate a 4-layer deep neural network with 40 neurons per hidden layer and randomly sample $N_{u_p}=100$ corresponding training data points, $N_p=600$ interior labeled data, $N_f=2000$ collocation points by Latin hypercube sampling from the training domain to predict the unknown parameter $\lambda$ as well as the numerical solution.

Under the same network structure and the same number of training points, Table \ref{tab5} shows the results of five randomized experiments for PINN and sdPINN, where the PINN is trained in the whole domain but the $L_2$ relative errors of the predicted solution are only tested in the same sub-domain as the sdPINN for a fair comparison. Obviously, the $L_2$ relative errors of sdPINN reach $10^{-3}$ orders of magnitude and the one of PINN is only $10^{-1}$ order of magnitude. When comparing the absolute error of calculating $\lambda$ by the PINN and sdPINN, the sdPINN clearly outperforms PINN by three orders of magnitude.
\begin{table}[htp]\scriptsize
\captionsetup{width=.9\textwidth,font={scriptsize}}
\caption{Case of $b=5$: Comparisons of $L_2$ relative errors and relative errors in $\lambda$ for different seeds. The training data are chosen as $N_u=100$, $N_p=600$ and $N_f=2000$.}
\centering
\renewcommand{\arraystretch}{1.2}
\begin{tabular}{c c| ccc c ccc}
\hline
           &&1&2&3&4&5&Mean \\ \hline
PINN       &$L_2$ relative error&{4.8013e-03}&{6.8694e-04}&{4.9065e-02}&{1.0033e+00}&{5.6071e-02}&{2.2279e-01}\\
           &Relative error in $\lambda$&{3.3219e+00}&{2.3038e-01}&{9.8086e+01}&{5.8937e+01}&{9.7945e+01}&{5.1704e+01}\\\hline
sdPINN     &$L_2$ relative error&{1.3905e-03}&{2.6572e-03}&{2.1930e-03}&{9.5490e-04}&{2.8050e-03}&{2.0001e-03}\\
           &Relative error in $\lambda$&{6.9737e-04}&{8.5109e-02}&{5.5227e-02}&{1.6518e-03}&{7.2655e-02}&{4.3068e-02}\\\hline
\end{tabular}
\label{tab5}
\end{table}

\begin{figure}[htp]
    \begin{minipage}{0.5\linewidth}
		\vspace{3pt}
		\centerline{\includegraphics[width=\textwidth]{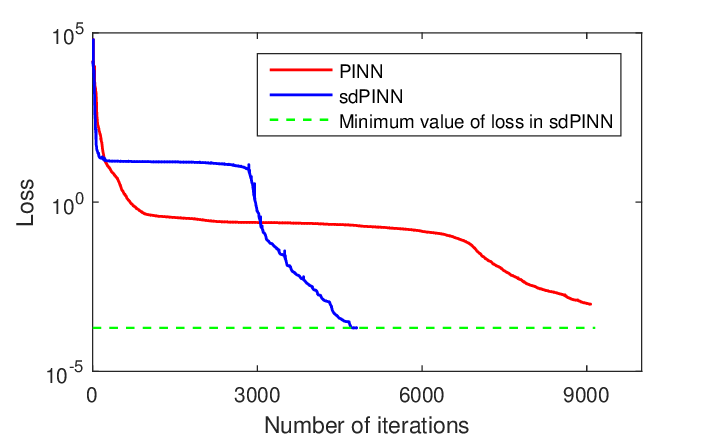}}
        \centerline{(A)}
	\end{minipage}
    \begin{minipage}{0.5\linewidth}
		\vspace{3pt}
		\centerline{\includegraphics[width=\textwidth]{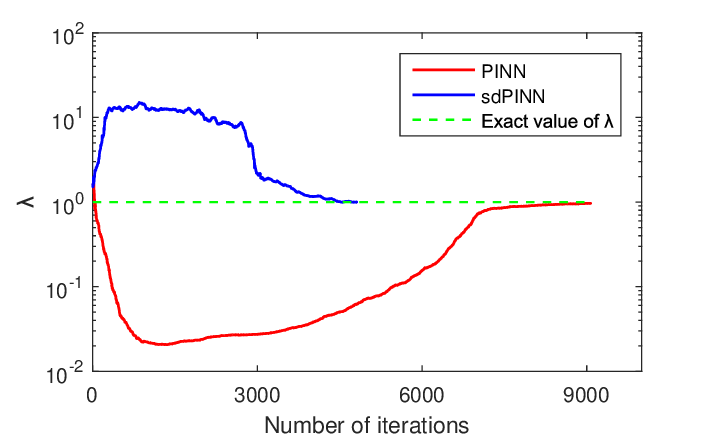}}
        \centerline{(B)}
	\end{minipage}
    \caption{(Color online) Case of $b=5$: The loss decrease tendencies and approximate tendencies of $\lambda$ by PINN and sdPINN. (A)Loss decrease. (B)Approximate tendencies of $\lambda$.}
\label{fig12}
\end{figure}

Furthermore, the decreasing tendencies of loss functions in Figure \ref{fig12}(A) shows that the red line for PINN expresses a slow decrease and arrives at the minima $9.6247\times10^{-4}$ with approximate 9000 iterations, but still bigger than the minima by sdPINN which reaches $1.9256\times10^{-4}$ and only requires $4809$ iterations. In Figure \ref{fig12}(B), the values of $\lambda$ by sdPINN rapidly approach the true value after $4300$ iterations, with a final prediction $1.0007$, but the $\lambda$ by PINN arrives at the true value after more than $9000$ iterations, with a final prediction of $0.9668$.

\subsubsection{Case of $b=20$}
In this case, we adopt the same architecture as $b=5$, but randomly sample $N_u=100$ training points, $N_p=500$ interior labeled data, $N_f=1000$ collocation points from the training domain to predict the parameter $\lambda$.

Similarly, we perform five independent trials to check the effects of PINN and sdPINN and display the results in Table \ref{tab6}. The PINN shows a poor performance for $b=20$ while the sdPINN gives a satisfactory result, where the mean $L_2$ relative error of the predicted solution is $5.1621\times10^{-3}$ and the mean absolute error of $\lambda$ is $1.3389\times10^{-2}$.
\begin{table}[htp]\scriptsize
\captionsetup{width=.9\textwidth,font={scriptsize}}
\caption{Case of $b=20$: Comparisons of $L_2$ relative errors of predicted solutions and relative errors of $\lambda$ for different seeds. The numbers of training data are chosen as $N_u=100$, $N_p=500$ and $N_f=1000$ respectively.}
\centering
\renewcommand{\arraystretch}{1.2}
\begin{tabular}{c c| ccc c ccc}
\hline
           &&1&2&3&4&5&Mean \\ \hline
PINN       &$L_2$ relative error&{1.0586e+00}&{1.5586e-03}&{1.0001e+00}&{4.5959e-01}&{1.0195e+00}&{7.0787e-01}\\
           &Relative error in $\lambda$&{6.5415e-01}&{1.0884e-03}&{5.0832e-01}&{2.4408e+00}&{6.2280e-01}&{8.4544e-01}\\\hline
sdPINN     &$L_2$ relative error&{1.1549e-02}&{1.8312e-03}&{1.2289e-03}&{2.0651e-03}&{9.1366e-03}&{5.1621e-03}\\
           &Relative error in $\lambda$&{3.7471e-02}&{1.0031e-02}&{1.1826e-04}&{6.1079e-03}&{1.3218e-02}&{1.3389e-02}\\\hline
\end{tabular}
\label{tab6}
\end{table}

\begin{figure}[htp]
    \begin{minipage}{0.5\linewidth}
		\vspace{3pt}
		\centerline{\includegraphics[width=\textwidth]{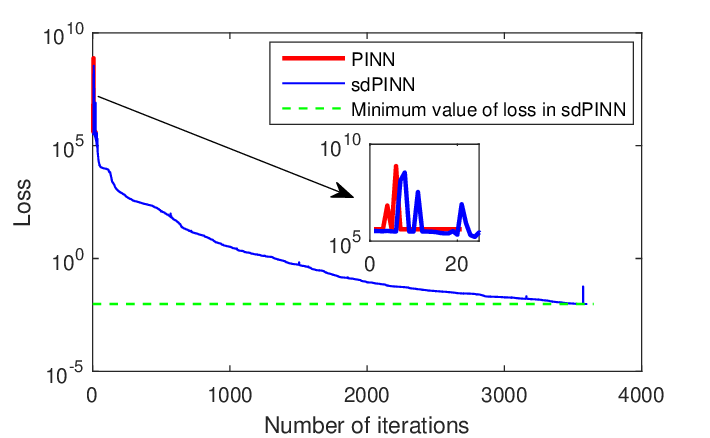}}
        \centerline{(A)}
	\end{minipage}
    \begin{minipage}{0.5\linewidth}
		\vspace{3pt}
		\centerline{\includegraphics[width=\textwidth]{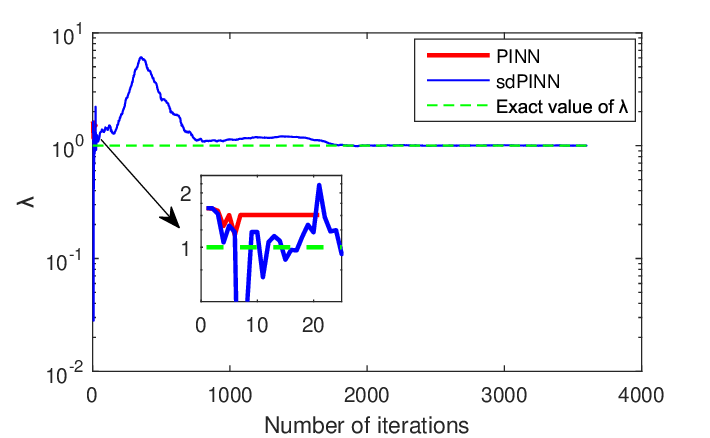}}
        \centerline{(B)}
	\end{minipage}
    \caption{(Color online) Case of $b=20$: The loss decrease tendencies and approximate tendencies of $\lambda$ by PINN and sdPINN. (A)Loss decrease. (B)Approximate tendencies of $\lambda$.}
\label{fig13}
\end{figure}

In Figure \ref{fig13}(A), we show the decrease trends of loss functions by PINN and sdPINN, where the red line for PINN stops at only $21$ iterations and the final value is $4.0619\times10^5$, while the blue line for sdPINN displays a rapid decrease and arrives at $9.5672\times10^{-3}$ after $4809$ iterations. Correspondingly, the predicted parameter $\lambda$ by PINN is $1.5083$ which is far from the true value, while the one by sdPINN closely approaches one and arrives at $0.9999$ after $3600$ iterations.

\section{Conclusion}
We propose a symmetry-group based domain decomposition strategy to enhance the power of PINN for finding high-accuracy numerical solutions of PDEs and discovering parameters in PDEs. Instead of solving the PDEs by one large PINN optimization problem, the idea is to use an approach of ``divide and conquer" and train many smaller optimization problems with parallel computations in the non-overlapping sub-domains. Such operations can reduce training times and potentially reduce the difficulties of the global optimization problem and also bring more flexibility in the training. Furthermore, in spite of the partition of computational domain, the proposed method still provides a continuous solution for the entire spatiotemporal domain which removes the emergence of the discontinuity problem of the learned solutions of the sub-domain.
On the other hand, with the introduced method, data-driven discovery of PDEs can be performed only in a sub-domain and does not need the prior labeled data, which largely reduces the computational cost of generating the labeled data in the mode of PINN.

However, there still exist some problems to be solved. One is how to divide the entire domain into a proper number of sub-domains which makes the neural network not only get high-accuracy solution but also take low computational cost. Maybe the self-adaptive strategy is an alternative way to give a reasonable partition of the domain. Another one is the deep integration of physical laws of PDEs with the neural networks. In addition to the aforementioned invariant surface condition induced by the Lie symmetry, the energy and momentum conservations generated by the conservation law of PDEs are also beneficial to enhance the learning ability of neural networks. Such topics are under consideration and will be reported in our future work.
\section*{Acknowledgements}
The paper is supported by the Beijing Natural Science Foundation (No. 1222014) and the Cross Research Project for Minzu University of China (No. 2021JCXK04); R\&D Program of Beijing Municipal Education Commission (Nos. KM202110009006 and KM201910009001).
\\\\
\textbf{Declarations of interest: none}

\end{document}